\documentclass{article}
\usepackage[margin=1in]{geometry}
\usepackage{natbib}
\usepackage{hyperref}

\usepackage{amsmath,amsthm,amsfonts,amssymb,bbm}
\usepackage{thmtools,booktabs}
\usepackage{graphicx}

\newtheorem{lemma}{Lemma}

\title{MolLedger: An Additive Graph Neural Network with Chemically Grounded ADME Attributions}

\author{Christina X. Ji\\
Hamilton College\\
\texttt{cji@hamilton.edu}
}

\begin{document}

\maketitle

\begin{abstract}
Optimizing absorption, distribution, metabolism, and excretion (ADME) is an important part of small molecule drug discovery. Many machine learning models have been built to predict ADME properties to facilitate this optimization process, but explaining model predictions is challenging. We propose a new graph neural network architecture with built-in atom attributions. Our model MolLedger learns a global context vector for each molecule and a per-atom head to output atom scores that sum to the predicted property. The atom scores are regularized to align with relevant chemical properties. We prove that MolLedger is a universal approximator and demonstrate empirically that the new architecture obtains explainability with little effect on performance. We show that the interpretations from MolLedger are faithful, concordant with held-out physical properties, and align with the changes between matched molecular pairs. Our case studies comparing interpretations from multiple methods on molecular pairs reveal that MolLedger is much better at producing sensible explanations for predicted property changes.
\end{abstract}

\section{Introduction}

\label{sec:intro}

During the lead optimization phase in drug discovery, chemists spend a significant amount of time changing subgroups in promising candidate drug molecules to improve properties related to absorption, distribution, metabolism, and excretion (ADME) \citep{jorgensen2009efficient}. Machine learning (ML) models can assist in this process by predicting these properties \citep{wu2018moleculenet,di2023systematic,rich2024machine}. Instead of synthesizing each compound and assaying these properties in a lab, chemists can run large virtual screens using computational models to surface the top drug candidates for further exploration. Traditional cheminformatics models predict ADME properties from molecular descriptors, such as partial charges, molecular weight, heavy atom count, connectivity indices, van der Waals surface area, polarity, and functional group counts \citep{ghose1986atomic,galvez1994charge,wildman1999prediction,ertl2000fast,consonni2009molecular}. ML models, such as random forests, gradient boosted trees, and neural networks, leverage these features as well. Graph neural networks (GNNs) are also popular for ADME prediction as they leverage the molecular structure \citep{liu2019chemi}. Nodes in the graph carry atom features, while edges carry bond features. Multi-task GNNs output predictions for multiple properties from a shared representation \citep{du2023admet,gu2024admetsar3}. However, these predictions may not always align with chemists' intuition. Making the ML model interpretable provides valuable guidance for medicinal chemists as they decide whether to trust model predictions and which molecules to select from virtual screens \citep{jimenez2020drug,wei2022interpretable,rich2024machine,ito2025improving}.

Previous works have tackled interpretability for ADME predictions with post-hoc, attention-based, and additivity-constrained approaches. Post-hoc interpretability methods, such as LIME, Grad-CAM, integrated gradients, or occlusion, use gradients or model-agnostic probes to examine how changes in each atom affect the overall prediction \citep{jimenez2020drug,wei2022interpretable,rao2022quantitative,jamrozik2024admet,ito2025improving,janssen2026machine}. Attention-based methods aggregate the amount of attention each atom is given across all the GNN layers to interpret how each atom contributes to the prediction \citep{tang2020self,guo2022ligandformer}. While post-hoc and attention-based approaches can compute the relevance of each atom to the prediction, we find that they may not reflect the exact value an atom contributes to the predicted label or align with physical properties. Another common approach to interpretability is to make the model a generalized additive model (GAM) \citep{hastie1986generalized}. \citet{bechler2406intelligible} apply this principle to GNNs and build a graph neural additive network (GNAN) that replaces  message passing with topology-based weighting of functions of individual features. GNAN produces faithful atom attributions that may not align with physical properties at the expense of performance. 

We build a GNN called MolLedger that provides faithful chemically grounded atom-level attributions for ADME predictions without sacrificing performance. Like GNAN, the prediction is  the sum of atom-specific scores, but the atom scores from MolLedger are computed with message passing and supported by a global context vector so that MolLedger is a universal approximator. These atom scores are also anchored to physical properties for meaningful interpretations. We demonstrate MolLedger achieves similar predictive performance to existing models and produces atom-level attributions that align better to chemical properties than existing interpretability methods. Code is available at \url{https://github.com/cxji/molledger}.

\section{MolLedger}

\label{sec:molledger_design}

\begin{figure}
	\centering
	\includegraphics[width=\linewidth]{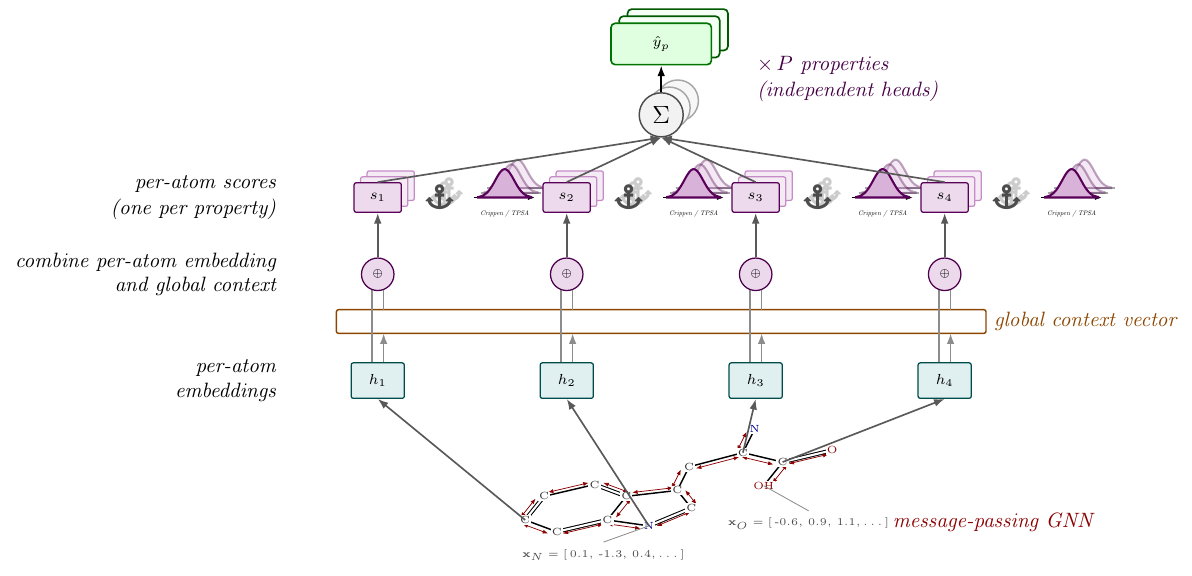}
	\caption{MolLedger has two novel components: 1) an additive head that uses a global context vector to generate per-atom scores and 2) an auxiliary loss to anchor the per-atom scores to a scaled version of a physical property.}
	\label{fig:schematic}
\end{figure}

MolLedger introduces two novel components: 1) A global context head that feeds into a per-atom head to generate scores for each atom that are then summed to the prediction. 2) An auxiliary loss that anchors per-atom scores to physical properties to ensure that the atom scores are meaningful. These two components are shown in Figure~\ref{fig:schematic}.

\subsection{Additive Atom Scores with Global Context}

Intelligible models like GAMs often sacrifice function capacity for interpretability. A two-layer multi-layer perceptron (MLP)  $\hat{y} = \psi_2 \left(\sum_j \psi_1 \left(x_j\right)\right)$ is a universal approximator \citep{hornik1989multilayer,hornik1991approximation}, but the black box cannot be interpreted. In contrast, GAMs cannot approximate as large of a class of functions because they lack the $\psi_2$ term \citep{agarwal2021neural,doohan2026comparison,chang2021node}, but the coefficients are intelligible. We can resolve the tension between function capacity and intelligibility when we note that our goal is per-atom attribution rather than per-feature attribution. Our model MolLedger is constrained to the form $\hat{y} = \sum_{i=1}^{N_{atoms}} \hat{s}_i$, where $\hat{s}_i$ can be a function of the features for all atoms rather than just the features for atom $i$. MolLedger performs message passing to learn a representation for each atom. Then, MolLedger produces a global context vector from the pooled embedding with its global context head $\phi_2$. Finally, MolLedger passes the global context vector and each atom's representation into a per-atom head $\phi_3$ to produce each $\hat{s}_i$. Mathematically, MolLedger can be written as:
\begin{equation}
    \hat{y} = \sum_{i=1}^{N_{atoms}} \hat{s}_i = \sum_{i=1}^{N_{atoms}} \phi_3 \left(A_i, \phi_2\left(\sum_{j=1}^{N_{atoms}} A_j, \frac{1}{N_{atoms}} \sum_{j=1}^{N_{atoms}} A_j\right)\right)
\end{equation}
where $A_i$ is the activation for atom $i$ from message passing. $\phi_2$ and $\phi_3$ are MLPs. Model details are in Appendix~\ref{app:details}. With this formulation, MolLedger is a universal approximator:
\begin{restatable}{theorem}{uniapproxthm}
    \label{thm:uni_approx}
    Let $\phi_1$ be the last MLP in the final message-passing layer that produces $A_i$. Let $H$ be the width of the input into $\phi_1$. Assume $H \ge N_{max} + 2$, where $N_{max} = \max N_{atoms}$ across all possible input molecules. The input into $\phi_1$ is a multiset $X$ of $n \le N_{max}$ atoms. $X$ can take on a finite set of values $\mathcal{X}$.
    
    Let $f$ be any function of a multiset of atoms and $u\left(x\right) > 0$ be any positive weight for each atom $x$. There exists parameters for $\phi_1$, $\phi_2$, and $\phi_3$ with any global context vector dimension $d \ge 1$ such that for every multiset $X$
    \begin{equation}
        s_x = u\left(x\right) \frac{f\left(x\right)}{\sum_{y \in \mathcal{X}} u\left(y\right)} \, \text{ for each atom } x \in X, \; \text{ and so } \sum_{x \in X} s_x = f\left(X\right).
    \end{equation}
\end{restatable}
The proof for this theorem is in Appendix~\ref{app:proof}. Because MolLedger is a universal approximator, MolLedger can achieve interpretability without sacrificing performance unlike the GAMs typically used for intelligible models.

\subsection{Auxiliary Loss for Anchors}

As there are many ways for the atom scores to sum to the property, minimizing the mean squared error loss may not give sensible scores for each atom. To obtain informative attributions, we add an auxiliary loss that anchors each atom towards a relevant chemical score. The two properties we use as anchors are Crippen score and topological polar surface area (TPSA). Crippen estimates the lipophilicity of a heavy atom based on its element type, neighboring atoms, and hybridization state \citep{wildman1999prediction}. Because logD is an adjusted version of the sum of the Crippen scores and a hydrophobic molecule is more likely to be metabolized, logD and the two measurements of clearance in mouse and human liver cells are anchored to Crippen. Kinetic solubility, aqueous solubility, and the fraction unbound measured for plasma protein binding in mouse plasma, brain, and muscle are inversely correlated with lipophilicity, so these five tasks are anchored to the negative Crippen value. TPSA measures the surface area around polar atoms and their attached hydrogens \citep{ertl2000fast}. Higher TPSA implies higher polarity, which tends to impede passive diffusion into the cell and promote active transport out of a cell, so Caco2 A-to-B permeability is anchored to the negative TPSA score while Caco2 efflux ratio is anchored to the positive TPSA score.

During training, MolLedger applies an auxiliary loss between the atom scores and the anchor properties. Because these anchors are not on the same scale as the properties themselves, the atom score and the anchor are normalized in the regularization term in this per-molecule per-task loss:
\begin{align}
    \mathcal{L} &= \frac{1}{\sigma_y^2} \left(y - \hat{y}\right)^2 + \frac{\lambda_{anchor}}{N_{atoms}} \sum_{i=1}^{N_{atoms}} \left(\frac{\hat{s}_i - \bar{s}}{\sigma_s} - \frac{
\text{sign}_{anchor} \left(a_i - \bar{a}\right)}{\sigma_a}\right)^2 \\
&= \frac{1}{\sigma_y^2} \left(y - \hat{y}\right)^2 + 2 \lambda_{anchor} \left(1 - \text{sign}_{anchor} r\right)
\end{align}
where $a_i$ is the anchor for atom $i$ and $r$ is the Pearson correlation between the anchors and predicted scores. $\sigma_y^2$ computed on the training set corrects for label scale in a multi-task loss. $\text{sign}_{anchor}$ is $+1$ for properties that are positively correlated with the anchor and $-1$ for properties that are negatively correlated as specified above. The second line shows this regularization penalty maximizes the correlation. We tune $\lambda_{anchor} \in \left\{ 0.1, 0.3 \right\}$ and compare with the ablation $\lambda_{anchor} = 0$ to evaluate the effectiveness of this auxiliary loss. We discuss the connections between this regularization term and prior work regarding regularization terms that match attributions to target values and concept bottleneck models in Appendix~\ref{app:details}.

\section{Experiment Set-up}

\label{sec:experiments}

\subsection{Datasets}

\label{sec:datasets}

We compiled 11 ADME tasks from the following datasets: 1) ExpansionRx (7,608 unique molecules) \citep{castellanoslessons} with logD, kinetic solubility, intrinsic clearance measured in mouse and human liver models, Caco2 efflux ratio, Caco2 A-to-B permeability assay, and plasma protein binding in mouse plasma, brain, and muscle. 2) Mollipo (4,200 unique molecules) \citep{wu2018moleculenet,hu2020open} with logD. 3) Therapeutics data commons (1,930 unique molecules) \citep{huang2021therapeutics} with Caco2 A-to-B permeability assay and half life. 4) ESol (1,117 unique molecules) \citep{delaney2004esol,wu2018moleculenet} with aqueous solubility. The merged registry includes 14,539 unique molecules. We followed the procedure in \citet{wu2018moleculenet} to split the molecules. Molecules were grouped by their Murcko scaffolds \citep{bemis1996properties}. All molecules in the same scaffold were assigned to the same split, yielding roughly 80\% train, 10\% validation, and 10\% test. Labels were converted to the same scale. As input to the model, we provide 9 types of categorical atom features and 3 types of categorical bond features following the convention in Open Graph Benchmark \citep{hu2020open}. More details are in Appendix~\ref{app:details}.

\subsection{Performance Evaluation}

\label{sec:performance_eval_setup}

ADME models can be evaluated on how accurately they predict properties (MAE), how well they rank all the molecules (Spearman correlation), and how accurately they predict the change between pairs of molecules that reflect swaps medicinal chemists may perform during lead optimization. The latter can be evaluated with matched molecular pairs \citep{tamura2021interpretation,turk2017coupling,lumley2020derivation,janssen2026machine}. We evaluate the matched pair $\Delta$ MAE for molecules $A$ and $B$ as $\left| \left(y_A - y_B\right) - \left(\hat{y}_A - \hat{y}_B\right) \right|$. Details on matched pair construction are in Appendix~\ref{app:details}.

We compare MolLedger to a standard pooled GNN  to show that adding an increasingly large global context vector bridges the performance drop introduced by the additivity constraint. To evaluate the effect of anchors, we compare MolLedger to a model trained without the anchor loss. We also compare MolLedger to other established ADME models. In Section~\ref{sec:intro}, we introduced GNAN  \citep{bechler2406intelligible}, a GNN that is linear in feature space, and LigandFormer \citep{guo2022ligandformer}, an attention-based GNN. GNAN has the same mean squared error loss, so we add our regularization term to its objective to distinguish between the contributions of our more expressive additive framework and the anchor term. As \citet{kamuntavivcius2025benchmarking} found that models with molecular descriptors perform well on ADME tasks, we also evaluate two models with molecular descriptors: 1) a pooled GNN with 227 molecular descriptors from RDKit \citep{rdkit} concatenated to the pooled graph embedding and 2) a gradient boosted tree (GBT) built with only the 227 molecular descriptors.

\subsection{Interpretability Metrics}

\label{sec:interp_setup}

To evaluate whether the atom scores MolLedger produces would be helpful to a medicinal chemist, we consider four axes that reflect how well the scores align to chemists' intuition, explain the model's predictions, localize changes between matched molecular pairs to swapped fragments, and match measured differences between pairs. We define each criterion mathematically below.

\textit{Concordance} Does the atom-level attribution correlate with known per-atom chemical ground truth? This is often called plausibility in explainable AI \citep{nauta2023anecdotal}. For example, hydrophobic atoms tend to contribute to higher logD scores, so a well-aligned attribution would give hydrophobic atoms higher scores than hydrophilic atoms. We measure this alignment with concordance:
\begin{equation}
    \text{Concordance} = \frac{1}{N_0 N_1} \sum_{i: p_i = 0} \sum_{j: p_j = 1} \text{sign}_p \mathbbm{1}\left\{s_i < s_j\right\} = AUROC(p, s)
\end{equation}
where $p$ is the binary property for each atom and $\text{sign}_p$ is the correlation direction. We evaluate concordance with hydrophobicity for logD, aromaticity for plasma protein binding in mouse plasma, hydrogen bond donor (HBD) identity for Caco2 A-to-B permeability assay, and hydrogen bond acceptor (HBA) identity for Caco2 efflux ratio. $\text{sign}_p = +1$ for logD and Caco2 efflux ratio and $-1$ for the other two tasks. Evaluating concordance with these properties that cannot be explained by the anchor shows that the model is learning chemical intuition rather than memorizing the anchor.

\textit{Exactness} Do the atom attributions sum to the prediction? This is known as faithfulness \citep{nauta2023anecdotal}. We report this as the MAE between the sum of the attributions and the predicted label:
\begin{equation}
    \text{Exactness gap} = \left| \hat{y} - \sum_{i=1}^{N_{atoms}} s_i\right|
\end{equation}
This is 0 by construction for both MolLedger and GNAN.

\textit{Localization} Given a matched pair, what proportion of the difference between their predicted scores is attributed to the swapped fragment rather than the common core? With $\mathcal{S}_A$ and $\mathcal{S}_B$ as the sets of atoms in the swapped fragments of molecules $A$ and $B$ in a matched pair and $\mathcal{S}_C$ as the set of atoms in the core, leakage to the core can be defined as
\begin{equation}
    \text{Leakage} = \frac{\left| \sum_{i \in \mathcal{S}_C} s_i^A - \sum_{i \in \mathcal{S}_C} s_i^B \right|}{\left| \sum_{i \in \mathcal{S}_C} s_i^A - \sum_{i \in \mathcal{S}_C} s_i^B \right| + \left| \sum_{i \in \mathcal{S}_A} s_i^A - \sum_{i \in \mathcal{S}_B} s_i^B \right|}
    \label{eq:leakage_def}
\end{equation}
\citet{janssen2026machine} also measure leakage, but they report the change in scores across the common core, the numerator in our metric. We normalize by the change across both groups to account for molecule size and property scale. Our definition of the core is given in Appendix~\ref{app:details}.

\textit{Explained $\Delta$ MAE} Given a matched pair, does the change on the substituent match the measured change in property? With $y_A$ and $y_B$ as the measured labels for molecules $A$ and $B$, the error in the explanation for each molecule is
\begin{equation}
    \text{Explained } \Delta \text{ absolute error} = \left| \left(\sum_{i \in \mathcal{S}_A} s_i^A - \sum_{i \in \mathcal{S}_B} s_i^B \right) - \left(y_A - y_B\right) \right|
\end{equation}

We evaluate these metrics on the atom scores from anchored and unanchored versions of MolLedger and GNAN. We also evaluate the attention-based interpretations from LigandFormer \citep{guo2022ligandformer} and apply integrated gradients (IG), LIME, Grad-CAM, and occlusion for molecules (WISP) to the pooled GNN to compare with other ADME interpretability work \citep{ito2025improving,jamrozik2024admet,wei2022interpretable,janssen2026machine}. Details on how these baselines are applied to the pooled GNN and evaluated with the interpretability metrics are in Appendix~\ref{app:interp_baselines}.

\section{Results}

\subsection{Model Performance}

\label{sec:perf_eval_results}

\begin{figure}
	\centering
	\includegraphics[width=\linewidth]{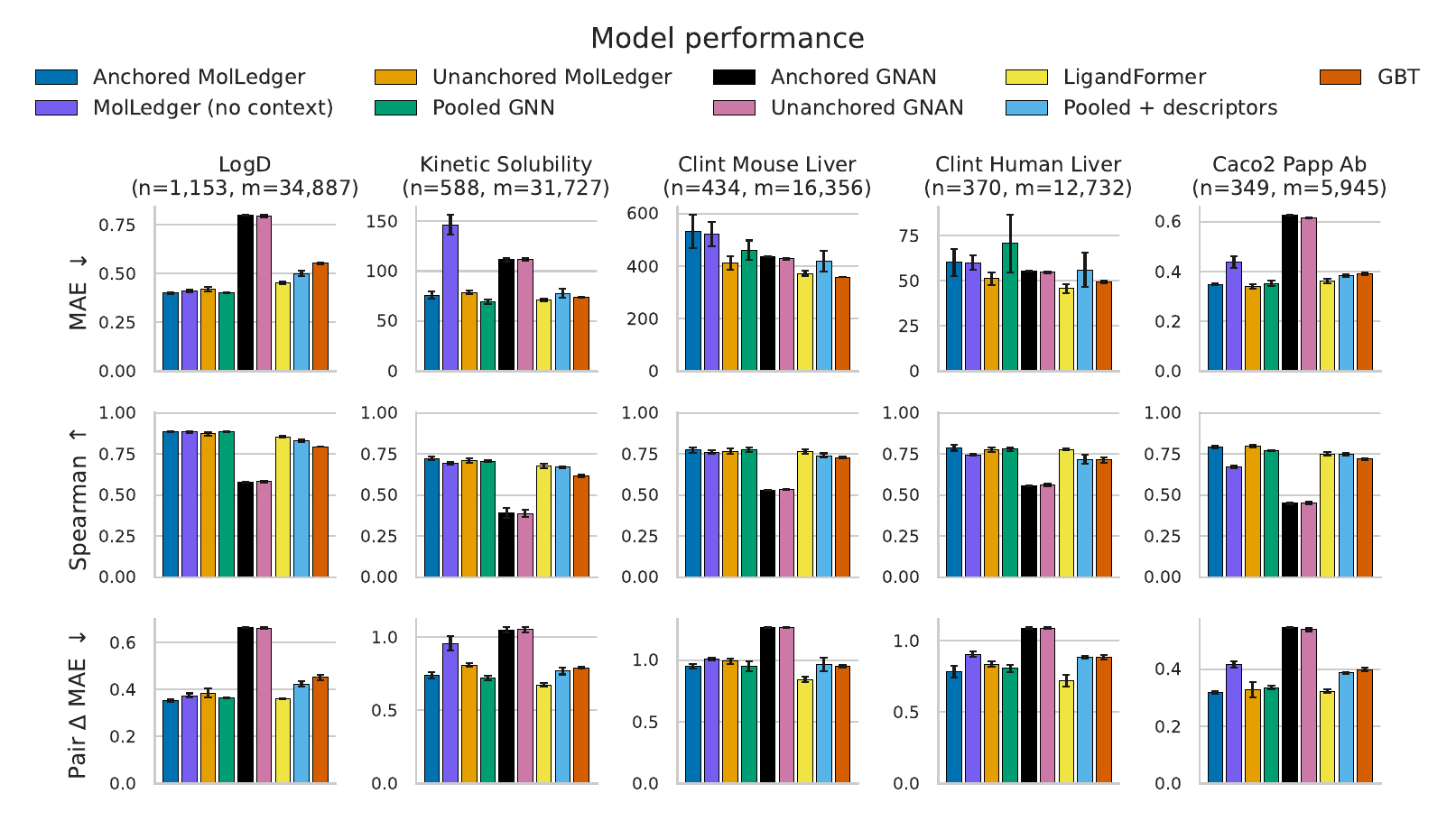}
	\caption{Performance on the held-out test set for the 5 largest tasks. $n$: number of test samples. $m$: number of matched pairs with a test molecule. All plots show mean and standard deviation across 3 random seeds. Remaining tasks in Figure~\ref{fig:performance_other} in Appendix~\ref{app:more_results}.}
	\label{fig:performance}
\end{figure}

Figure~\ref{fig:performance} shows the results of the performance evaluation outlined in Section~\ref{sec:performance_eval_setup} on the 5 largest tasks, with results for additional tasks in Figure~\ref{fig:performance_other} and Tables~\ref{tab:perf-mae}-\ref{tab:perf-matchedpair} in Appendix~\ref{app:more_results}. Overall, MolLedger and the baselines perform similarly on MAE, Spearman, and matched-pair $\Delta$ MAE. The additive design of MolLedger with the global context vector does not have a negative effect on performance compared to the pooled GNN. This aligns with our theoretical results that MolLedger is a universal approximator. In contrast, GNAN and MolLedger without the global context vector perform worse on kinetic solubility and Caco2 A-to-B  permeability, and GNAN also performs worse on logD. This aligns with how they are not universal approximators.

Introducing the anchor loss in MolLedger has a mixed effect on performance. The results can be explained in part by anchor relevance to the task. For instance, Crippen is fairly relevant to LogD ($r=0.50$ in Table~\ref{tab:anchor-relevance}) as LogD is the sum of Crippen scores with an adjustment for ionization state, so anchoring has little effect on performance for logD. In contrast, Crippen is only minimally relevant to the two clearance tasks ($r=0.15$ and $r=0.12$) as hydrophobicity only partly explains metabolic stability. MolLedger performance deteriorates for the two clearance tasks with anchors. However, relevance alone does not explain the effect of anchoring. Crippen is highly relevant to aqueous solubility ($r=0.83$), yet MAE gets slightly worse because the unanchored model already somewhat agrees with the Crippen score. Furthermore, the unanchored model has more flexibility to account for the melting point term in the aqueous solubility equation that Crippen does not account for. In contrast, the anchor is not relevant to kinetic solubility ($r=0.17$) and Caco2 A-to-B permeability ($r = 0.36$), yet anchoring MolLedger does not hurt performance for those tasks. These changes are observed in MAE. Spearman and matched pair $\Delta$ MAE change very little with anchoring.

\subsection{Evaluation of Atom-level Attributions}

\label{sec:interp_eval_results}

\begin{figure}[t]
	\centering
	\includegraphics[width=\linewidth]{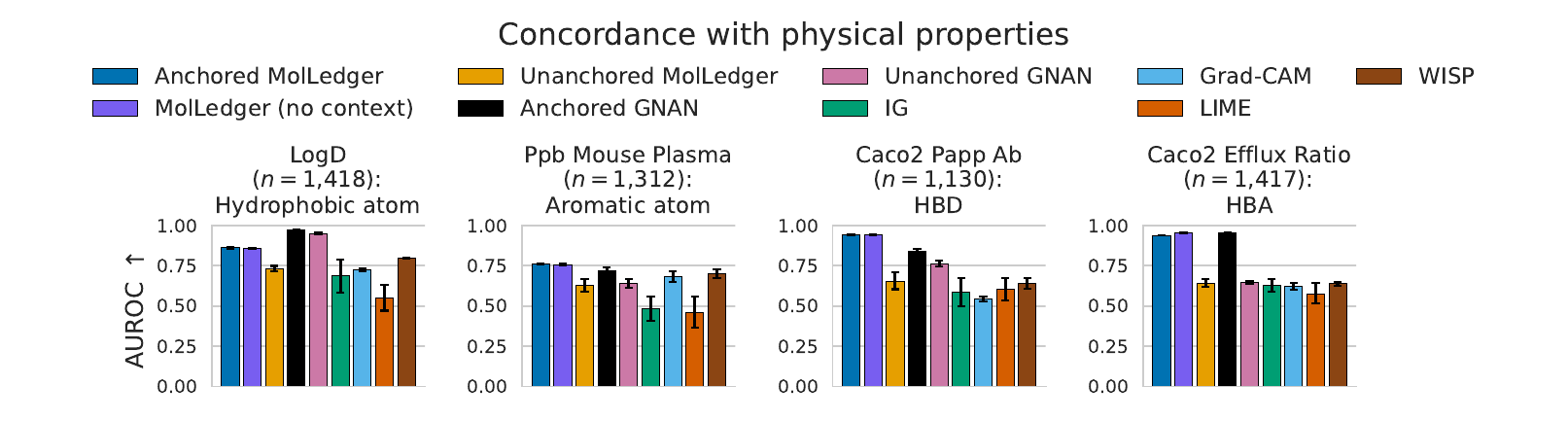}
	\caption{Concordance on the test set to relevant physical properties that differ from the anchors. As no labels are required, concordance is evaluated on all molecules in the test set that have at least 1 atom in each binary class (e.g. at least 1 HBD and 1 non-HBD).}
	\label{fig:noncircular_concordance}
\end{figure}

Figure~\ref{fig:noncircular_concordance} shows that anchoring MolLedger significantly improves concordance with physical properties that are distinct from the anchor. The anchored models identify HBAs are important for the Seelig motif in the P-gp active transport pump \citep{seelig1998general} and place higher scores on HBAs when predicting Caco2 efflux ratio. The anchored models also identify the presence of HBDs impedes passive diffusion into a cell and place low scores on HBDs for Caco2 A-to-B permeability. MolLedger and some other methods have similar concordance on LogD and plasma protein binding (PPB). This may be because hydrophobicity and aromaticity are only weak predictors for these tasks as molecules have many hydrophobic and aromatic atoms that may not all be relevant and thus do not actually need to have high scores for LogD or low scores for PPB.

\begin{figure}
	\centering
	\includegraphics[width=\linewidth]{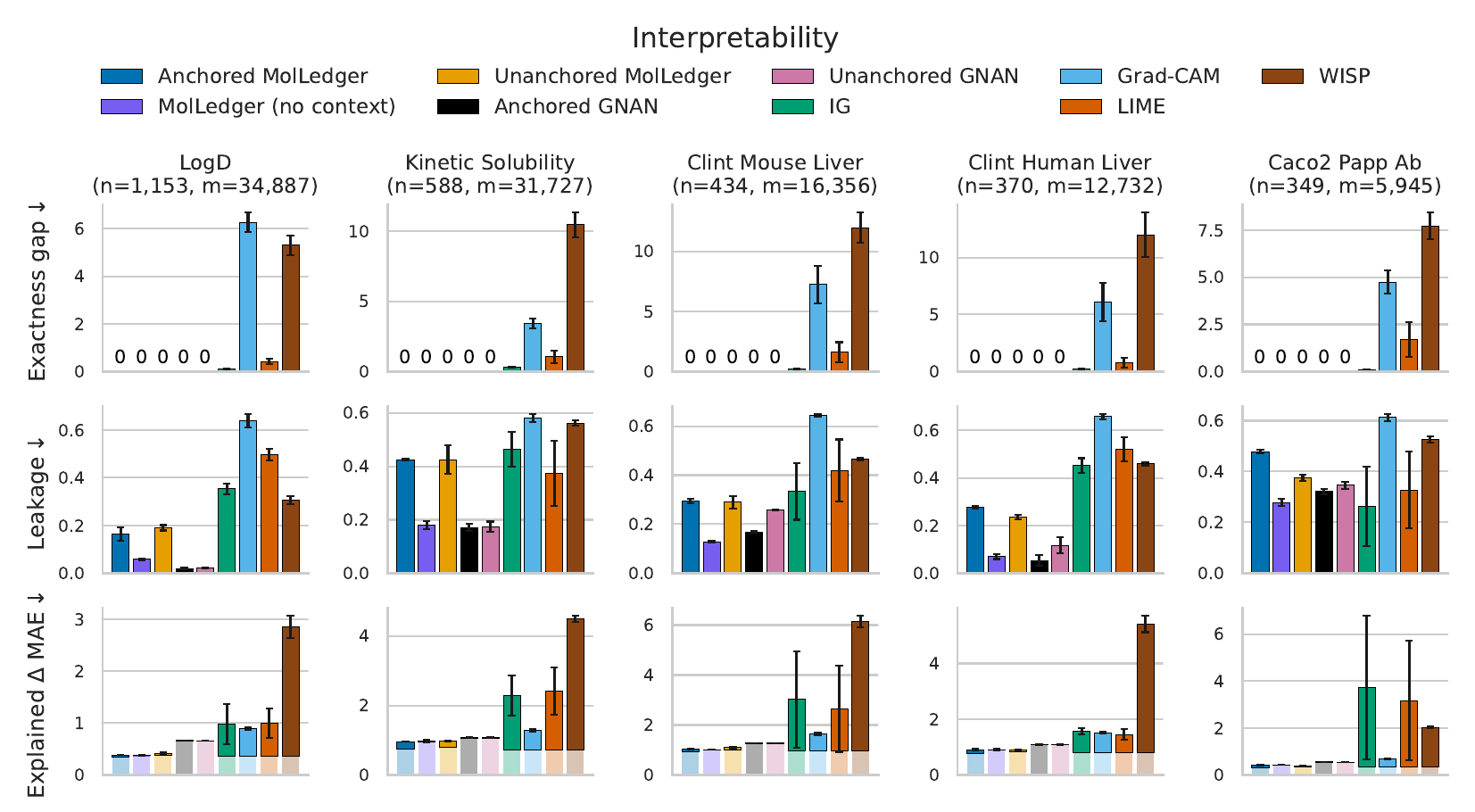}
	\caption{Interpretability metrics on test set for the 5 largest tasks. $n$ and $m$ same as Figure~
    \ref{fig:performance}. The transparent part of the explained $\Delta$ MAE bars is the MAE for the predicted differences shown in Figure~\ref{fig:performance}. The solid part represents attribution error. Remaining tasks in Figure~\ref{fig:interpretability_other} in Appendix~\ref{app:more_results}.}
	\label{fig:interpretability}
\end{figure}

Figure~\ref{fig:interpretability} evaluates the other interpretability metrics defined in Section~\ref{sec:interp_setup} on the 5 largest tasks, with additional results for other tasks in Figure~\ref{fig:interpretability_other} and Tables~\ref{tab:noncircular-concordance}-\ref{tab:interp-explained} in Appendix~\ref{app:more_results}. First, we note that the exactness gap of 0 that is built into MolLedger and GNAN is not a property that other methods achieve. While LIME is close to approximating the prediction, Grad-CAM has an extremely large exactness gap because the gradient is only approximated at a single point and the baseline correction term that is estimated with all-zero activations is likely out of distribution. The exactness gap for WISP is large as its attributions are designed to reflect swaps from multiple baseline molecules and no baseline correction term can fix the sum of the attributions to match the prediction \citep{janssen2026machine}. The exactness gap for IG is small only because we give IG more than twice as much time as other interpretability methods besides WISP to compute its gradient steps. We discuss this trade-off shown in Figure~\ref{fig:ig_steps} more in Appendix~\ref{app:interp_baselines}. In general, Figure~\ref{fig:timing} in Appendix~\ref{app:interp_baselines} shows how MolLedger is much faster to run than all the other interpretability methods, GNAN, and LigandFormer. Grad-CAM is the only method with a similar runtime because its single-point gradient approximation is fairly lightweight, but its single-point approximation is not faithful.

The plots of leakage in Figure~\ref{fig:interpretability} show that holding the core constant and only changing scores on the substituents is challenging for all tasks besides logD. LogD is the only property that can almost be decomposed into a sum of per-atom scores. Other properties are affected by interactions across fragments and cannot be separated cleanly between the core and the substituent. MolLedger without the global context vector and GNAN, which does not have message-passing between atoms, show less leakage into the core, but the performance of these models is much lower. We explore this trade-off further in Figure~\ref{fig:gctx_dim} in Appendix~\ref{app:variants}, where we see that increasing the width of the global context vector also increases leakage until context is saturated. Our ablation results align with chemical intuition that the trade-off between leakage and performance may be unavoidable. Nonetheless, MolLedger still has less leakage compared to other interpretability methods on most tasks.

\begin{figure}
	\centering
	\includegraphics[width=.85\linewidth]{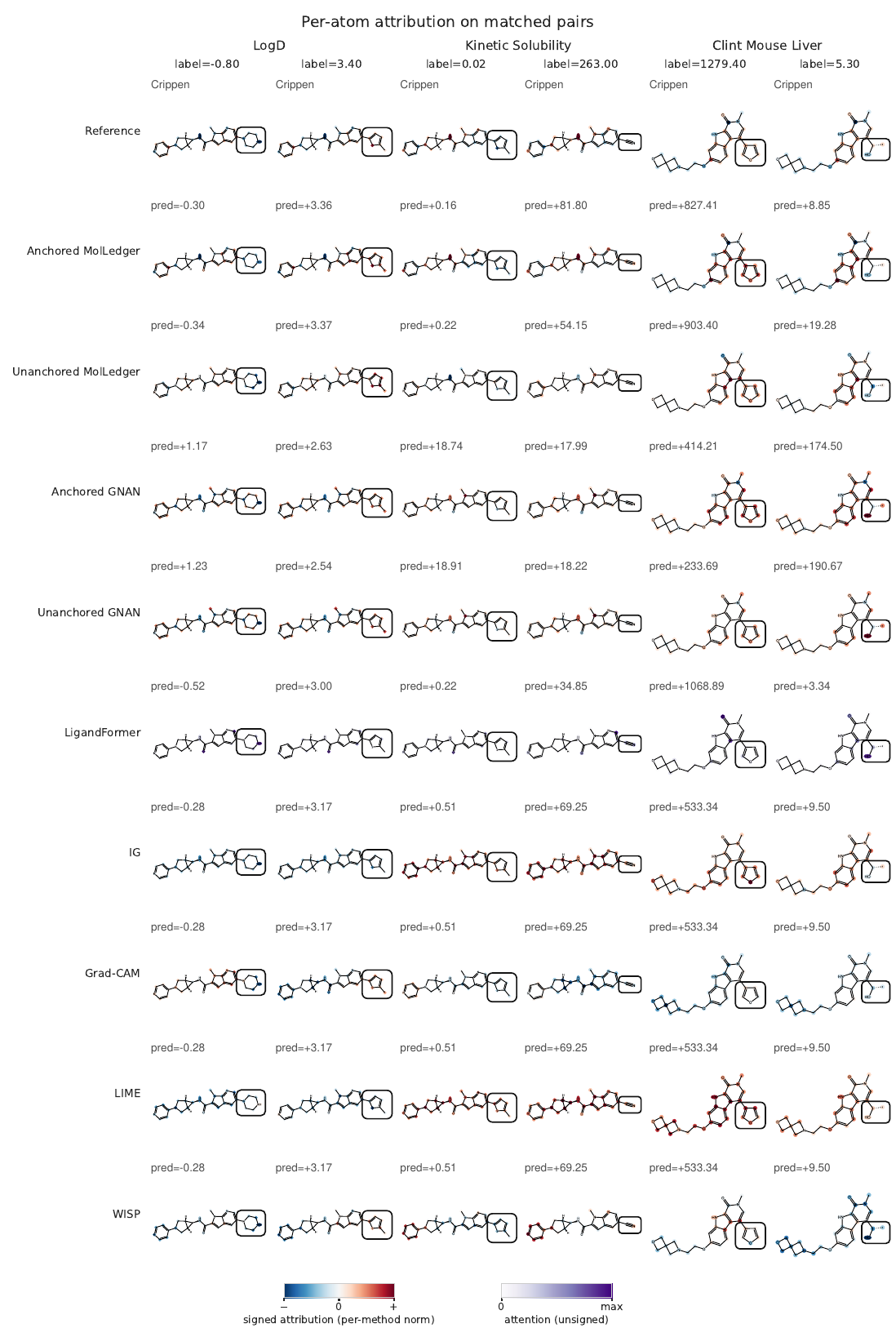}
	\caption{Per-atom attributions on matched molecular pairs with the largest property change for logD, kinetic solubility, and intrinsic clearance in mouse liver cells. Swapped fragments are shown in a box. The Crippen scores for each atom are shown in the top row for reference.}
	\label{fig:visual}
\end{figure}

The explained $\Delta$ MAE shown in Figure~\ref{fig:interpretability} synthesizes several metrics we have already seen. First, a model needs to be good at predicting pairwise differences for the explanation to be sensible. Then, if the attributions sum to the predicted value and are localized to the swapped fragment, the change in attributions in the swapped fragment approximates the predicted change, which is close to the measured change for a good model. Thus, explained $\Delta$ MAE is small when the $\Delta$ MAE evaluated in Figure~\ref{fig:performance}, exactness gap, and leakage are small. MolLedger has similar $\Delta$ MAE to other models, an exactness gap of 0, and relatively low leakage. Putting these factors together results in MolLedger having small explained $\Delta$ MAE on almost all tasks. In fact, most of the gap between the sum of the attributions and the measured difference is due to the prediction gap rather than attribution localization error. GNAN is worse because the model is poor at predicting pairwise differences. The pooled GNN has similar predictive performance to MolLedger, but the posthoc interpretability methods perform poorly because their attributions are incorrect. This is due to high leakage for all the posthoc methods and a large exactness gap for Grad-CAM and WISP. Combined with high concordance to physical properties, MolLedger's good performance on explained $\Delta$ MAE is a strong indicator that MolLedger can provide chemically grounded explanations to support predicted ADME changes during the lead optimization process.

\subsection{Case Studies}

\label{sec:case_studies}

Figure~\ref{fig:visual} shows the interpretations each method produces on matched molecular pairs with the largest difference in logD, kinetic solubility, and intrinsic clearance in mouse liver cells. In the example for logD, a hydrophilic piperazine in the left molecule is swapped for a hydrophobic methylthiophene. MolLedger correctly attributes negative contributions to atoms in the piperazine and positive contributions to atoms in the methylthiopene. Grad-CAM and WISP provide reasonable attributions as well though Grad-CAM incorrectly ascribes a positive score to a nitrogen in the piperazine. The interpretations from other methods do not align with chemical intuition. Because GNAN does not account for the identity of atom neighbors, GNAN cannot distinguish between carbons in a cyclohexane and carbons in a piperazine. Therefore, while GNAN correctly places negative scores on the polar nitrogens, it is unable to also place negative scores on the carbons in the piperazine. LigandFormer attends to a nitrogen in the piperazine but not to the methylthiophene that is swapped in. IG and LIME place negative scores on both fragments, which does not explain the change in prediction.

The molecular pair for kinetic solubility swaps the lipophilic aromatic ring in methylthiophene for a small polar nitrile. The lipophilic ring allows the former molecule to leave solution, while the small polar nitrile helps the latter compound stay in solution. Anchored MolLedger and WISP are the only methods that correctly place negative attributions on all atoms in the methylthiophene and positive attributions on both atoms in the nitrile. LigandFormer does not attend to the methylthiophene, and GNAN puts little weight on either  swapped fragment. Grad-CAM places negative attributions on atoms in both groups, while IG and LIME place positive attributions on atoms in both groups. The opposite attributions that different interpretability methods place on the same GNN demonstrate how posthoc interpretability methods are not reliable for understanding what drives model predictions.

In the example for intrinsic clearance in mouse liver cells, a furan is swapped for a hydroxyethyl. Furan is a metabolic soft spot because Cyp450 tends to epoxidize the ring, which is then opened via oxidation to form unstable intermediates. In the molecule on the right, the hydroxyl is polar and therefore reduces the likelihood of that molecule entering the hepatocyte and being cleared. Only MolLedger and IG explain both reasons for the decreased clearance. In contrast, LigandFormer and Grad-CAM do not attribute much weight to the furan, GNAN and LIME do not give the hydroxyl negative scores, and WISP incorrectly places a negative score on the oxygen in the furan. The only valid swap that WISP can compare against to compute the attribution on the oxygen in the furan is a swap from oxygen to sulfur and the resulting thiophene is also a metabolic soft spot. Additional examples for other tasks are shown in Figures~\ref{fig:visual_other} and~\ref{fig:visual_ppb}. Our analysis in Appendix~\ref{app:more_results} reflects on nuances, such as misleading anchors and the introduction of changes that interact with the core.

\section{Discussion}

MolLedger is the first ADME prediction model to our knowledge that both achieves good performance as a universal approximator and provides strong explanations for changes between matched molecular pairs. MolLedger attains this by adding a global context head and per-atom head to output atom scores from a GNN and applying a regularization term to anchor the atom scores to physical properties. Our experiments demonstrate that MolLedger performs on par with other ADME models, while its explanations are better than posthoc interpretability methods in terms of concordance with physical properties, localization to the swapped fragments in matched pairs, and faithfulness to the prediction and measured differences. Our case studies and evaluation results show that the atom attributions from MolLedger are meaningful explanations for how fragment swaps between matched molecular pairs change ADME properties. MolLedger is a promising candidate for adoption in virtual screens. The explanations will help chemists identify which molecules recommended by the model are worth pursuing in the lab and build trust in the model. The architectural contributions of MolLedger may also be applied to other graph-based tasks.

\section*{Acknowledgements}
We thank David Williams, Xiao Chen, and anonymous reviewers for feedback on a draft of this manuscript.

\bibliographystyle{plainnat}
\bibliography{reference}

\appendix

\section{Proof of Theorem~\ref{thm:uni_approx}}

\label{app:proof}

Before we prove Theorem~\ref{thm:uni_approx}, we give this lemma regarding the power sum of multisets:
\begin{lemma}
    \label{lem:power_sum_multiset}
    Let $t: \mathbb{R}^d \rightarrow \mathcal{R}$ be an injective function. For a multiset $X$ of $n \le N_{max}$ atoms, let $p_k\left(X\right) = \sum_{x \in X} t\left(x\right)^k$ for $k = 0, 1, \ldots, N_{max}$. The tuple $\left(p_0, \ldots, p_{N_{max}}\right)$ determines $X$.
\end{lemma}
\begin{proof}
    $p_0$ is the number of atoms $n$. $p_1$ to $p_n$ can be used to define the elementary symmetric polynomials $e_1, \ldots, e_n$ of the $n$ numbers $t\left(x\right)$ using Newton's identities. Vieta's formulas state that these elementary symmetric polynomials are the coefficients of the polynomial $\prod_{x \in X} \left(z - t\left(x\right)\right)$. A polynomial determines its roots, so the power sums determine $t\left(x\right)$. $t$ is an injective function, so the power sums determine $X$.
\end{proof}

Now we can use this lemma to prove Theorem~\ref{thm:uni_approx}, restated here:
\uniapproxthm*
\begin{proof}
    Because 2-layer MLPs are universal approximators, there exist parameters for $\phi_1$ such that 
    \begin{equation}
        \phi_1\left(x\right) = \left(t\left(x\right)^0, \ldots, t\left(x\right)^{N_max}, u\left(x\right), 0, \ldots, 0\right) \in \mathbb{R}^H
    \end{equation}
    Then, the pooled output is
    \begin{equation}
        \mathbf{P} = \sum_{x \in X} \mathbf{A}_x = \left(p_0\left(X\right), \ldots, p_{N_{max}}\left(X\right), U, 0, \ldots, 0\right)
    \end{equation}
    with $U = \sum_{y \in X} u\left(y\right)$.

    By Lemma~\ref{lem:power_sum_multiset}, the power sums in $\mathbf{P}$ determine $X$ and thus $f\left(X\right)$, so there exists some function $\rho$ such that $f\left(X\right) = \rho\left(\mathbf{P}\right)$. Again because 2-layer MLPs are universal approximators, there exists some parameters for $\phi_2$ such that we can map the pooled inputs $\left(\mathbf{P}, \mathbf{P}/n\right)$ to a size-$c$ global context vector $\mathbf{g}$ where the first dimension $g_0 = \rho\left(\mathbf{P}\right)/U$ and the other dimensions $g_1 = g_2 = \ldots = g_{c-1} = 0$.

    Now we will define the per-atom head such that $s_x = \phi_3\left(\mathbf{a}, \mathbf{g}\right) = u\left(x\right) g_0$. Because we defined $\phi_1$ such that $u\left(x\right)$ is an entry in $\mathbf{a}$, $\phi_3$ is a product of two of its inputs. By the universal approximation theorem for MLPs, there exist parameters for $\phi_3$ to define this function.

    Now we have shown the two desired results:
    \begin{equation}
    s_x = \phi_3\left(\mathbf{a}, \mathbf{g}\right) = u\left(x\right) g_0 = u\left(x\right) \frac{\rho\left(\mathbf{P}\right)}{U} = u\left(x\right) \frac{f\left(X\right)} {U}
    \end{equation}
    and 
    \begin{equation}
        \sum_{x \in X} s_x = \frac{f\left(X\right)}{U} \sum_{x \in X} u\left(x\right) = f\left(X\right).
    \end{equation}
\end{proof}

\section{Model and Dataset Details}

\label{app:details}

The underlying architecture of MolLedger is a GNN with 4 message-passing blocks. Message passing utilizes the GINEConv operator that takes in both node and edge features \citep{xu2018powerful,hu2019strategies}. Each message-passing block utilizes a 2-layer MLP that has 256 hidden units, applies batch normalization and ReLU, and outputs a 128-unit vector. The block that produces the global context and pooled GNN output is also a 2-layer MLP with 128 hidden units. It maps to a single scalar in the pooled GNN and a global context vector in MolLedger. Finally, the per-atom head is also a 2-layer MLP with 128 hidden units. All models were trained using an Adam optimizer using learning rate 1e-3 and cosine annealing schedule, batch size 64, 300 epochs, and 3 random seeds on a NVIDIA GeForce RTX 5070 Ti GPU.

When we merged the datasets from the sources outlined in Section~\ref{sec:datasets}, we converted the labels to the same scale. As Caco2 A-to-B was recorded in 1e-6 cm/s in ExpansionRx and $\log_{10}$ cm/s in Therapeutics data commons (TDC), we converted ExpansionRx to TDC's convention when merging. Kinetic solubility, Caco2 efflux ratio, half life, plasma protein fraction unbound, and clearance were converted to log scale. Human intestinal absorption was excluded as the labels were binary, and binary tasks do not fit with the additive anchored framework. Following Open Graph Benchmark's convention \citep{hu2020open}, the 9 types of categorical atom features  provided to the models are atomic number, chirality, degree, formal charge, total attached H-count, number of radical electrons, hybridization, aromaticity, and ring membership. We also provide 3 types of categorical bond features: bond type, stereochemistry, and conjugation.

Regarding the assignment of anchors to tasks, the sum of the Crippen scores of all the heavy atoms in a molecule is its partition coefficient, known as logP. Because logP assumes the molecule is neutral whereas many drug molecules are charged under physiological conditions, \textit{logD} is an adjusted version of logP that accounts for ionization. Even with this adjustment, logD is highly correlated with the sum of the Crippen scores, so we would expect Crippen to be an excellent anchor for logD. Hydrophobic molecules tend to have higher logD, so the hydrophobicity of individual atoms is used for the concordance evaluation. The two measurements of \textit{clearance in mouse and human liver cells} are anchored to Crippen as well. Hydrophobic molecules are more likely to be metabolized in the liver because they are more likely to cross the cell membrane and bind to the hydrophobic pocket in cytochrome P450. Both \textit{kinetic solubility} and \textit{aqueous solubility} are inversely correlated with lipophilicity, so we anchor the atom-level attributions for these two tasks to the negative Crippen value. \textit{Plasma protein binding in mouse plasma, brain, and muscle} measure fraction unbound. Because plasma proteins also have hydrophobic pockets, the fraction unbound label is also anchored to the negative Crippen value. Aromatic rings are particularly helpful for binding to plasma proteins, so aromatic atoms were used as the binary property for the concordance evaluation with mouse plasma chosen as the task to evaluate since it has the largest sample size.

As discussed in Section~\ref{sec:molledger_design}, high polarity tends to impede passive diffusion into a cell and promote recognition by the P-glycoprotein pump (P-gp) responsible for pumping a drug out of the endothelial Caco2 cells used in assays \citep{seelig1998general}. Therefore, \textit{Caco2 A-to-B permeability}, which measures influx rate, tends to be higher for less polar molecules and is anchored to the negative TPSA score, while \textit{Caco2 efflux ratio}, which measures the rate a drug is  pumped out relative to the influx rate, tends to be higher for more polar molecules and is anchored to the positive TPSA score. HBDs tend to impede passive diffusion, and HBAs are part of the Seelig motif that P-gp pumps recognize \citep{seelig1998general}, so those two properties were chosen for the concordance evaluation.

Half-life is the amount of time it takes for the concentration of a drug in blood or plasma to decrease by 50\%. Half-life is high when the molecule is distributed widely and cleared at a slow rate \citep{gibaldi2012pk}. Distribution and clearance are both faster when the molecule is hydrophobic, so half-life is not clearly correlated in either direction. Therefore, no anchor is applied for half-life. The relevance of the anchors to the other tasks are shown in Table~\ref{tab:anchor-relevance}.

Previous works have enforced attributions match targets in a variety of ways. \citet{ross2017right} regularizes the attribution for irrelevant features to 0. \cite{erion2021improving}, \citet{weinberger2020learning}, and \citet{rieger2020interpretations} apply a regularization term to match gradient-based attributions or decompositions that treat the model as a black box to known priors about target values and relationships between features. As the attributions $s_i$ that are regularized for MolLedger are intermediate outputs, $s_i$ is also analogous to the high-level concepts in concept bottleneck models \citep{koh2020concept}, where the concepts are atom properties in this case. Similar ideas have been applied to graph neural networks before \citep{xu2025graph,wang2025interpretable}. However, MolLedger is more constrained than concept bottleneck models. The function to predict the label from the concepts is defined to be the sum of the concepts rather than learned. This regularization reduces the space of solutions for the atom scores $\mathbf{s} = \left(s_0, \ldots, s_{n-1}\right) \in \mathbb{R}^n$ from an $\left(n-1\right)$-dimensional subspace $\sum_i s_i = y$ to a $1$-dimensional subspace $\mathbf{s} = \mathbf{1} * \frac{y}{n} + \gamma \left(\mathbf{a} - \mathbf{1} * \bar{a}\right)$, where $\gamma$ is the only remaining degree of freedom because we do not know the relative scale between the anchor and the atom score.

\begin{table}
\centering
\caption{Relevance of anchors used for each task. Relevance is the Pearson correlation between the sum of the anchors and the measured labels on the training set. The Pearson correlation is multiplied by -1 if the direction the anchor is applied is $-$.}
\label{tab:anchor-relevance}
\vspace{0.2cm}
\begin{tabular}{lrccc}
\toprule
Task & $n_{\text{train}}$ & Anchor & Direction & Relevance $r$ \\
\midrule
LogD & 9,233 & Crippen & $+$ & 0.50 \\
Kinetic solubility & 6,130 & Crippen & $-$ & 0.17 \\
CLint mouse & 4,810 & Crippen & $+$ & 0.15 \\
CLint human & 3,792 & Crippen & $+$ & 0.12 \\
Caco2 Papp Ab & 3,964 & TPSA & $-$ & 0.36 \\
Caco2 efflux & 3,235 & TPSA & $+$ & 0.36 \\
Aqueous solubility & 857 & Crippen & $-$ & 0.83 \\
PPB plasma & 1,561 & Crippen & $-$ & 0.48 \\
PPB brain & 1,286 & Crippen & $-$ & 0.45 \\
PPB muscle & 400 & Crippen & $-$ & 0.52 \\
\bottomrule
\end{tabular}
\end{table}

Predicting property differences between pairs is challenging because a small change can lead to a large property difference, known as an activity cliff \citep{stumpfe2014recent}. To construct matched pairs in our evaluation, we identify pairs of molecules that share a common core and have a single swapped side group similar to how chemists may refine drug candidates. We use the fragment-and-index algorithm implemented in rdkit \citep{rdkit, hussain2010computationally,dalke2018mmpdb}. The algorithm fragments each molecule by cutting on one acyclic, non-conjugated, single bond. If a single cut is sufficient to break the molecule into two fragments, the smaller fragment is considered the substituent and the larger fragment is considered the core. The substituent fragment may have up to 12 atoms. We include two additional cases that swap a single heavy atom with a hydrogen atom or another heavy atom connected by a double or triple bond (preserving bond order). At least one molecule in each matched pair must be in the test set.

In our evaluation of leakage and explained $\Delta$ MAE, the core is defined as the largest possible matched fragment that can be produced from our bond-cutting logic with the linker atom removed. The linker atom is the atom in this shared fragment that is bonded to the substituent group. Two molecules in a matched pair must have their fragments attached to the same linker atom. Since the neighbors of the linker atom have changed, it is reasonable to expect the attribution to the linker atom to change as well, similar to how Crippen score changes depending on an atom's neighbors. Changes to the score on the linker atom are included as part of the substituent change and not considered changes to the core. A good explanation associates the change in prediction to the substituent groups and the linker atom, not the common core.

\section{Tuning the Context Vector and Anchor Loss in MolLedger}

\label{app:variants}

\begin{figure}
	\centering
	\includegraphics[width=\linewidth]{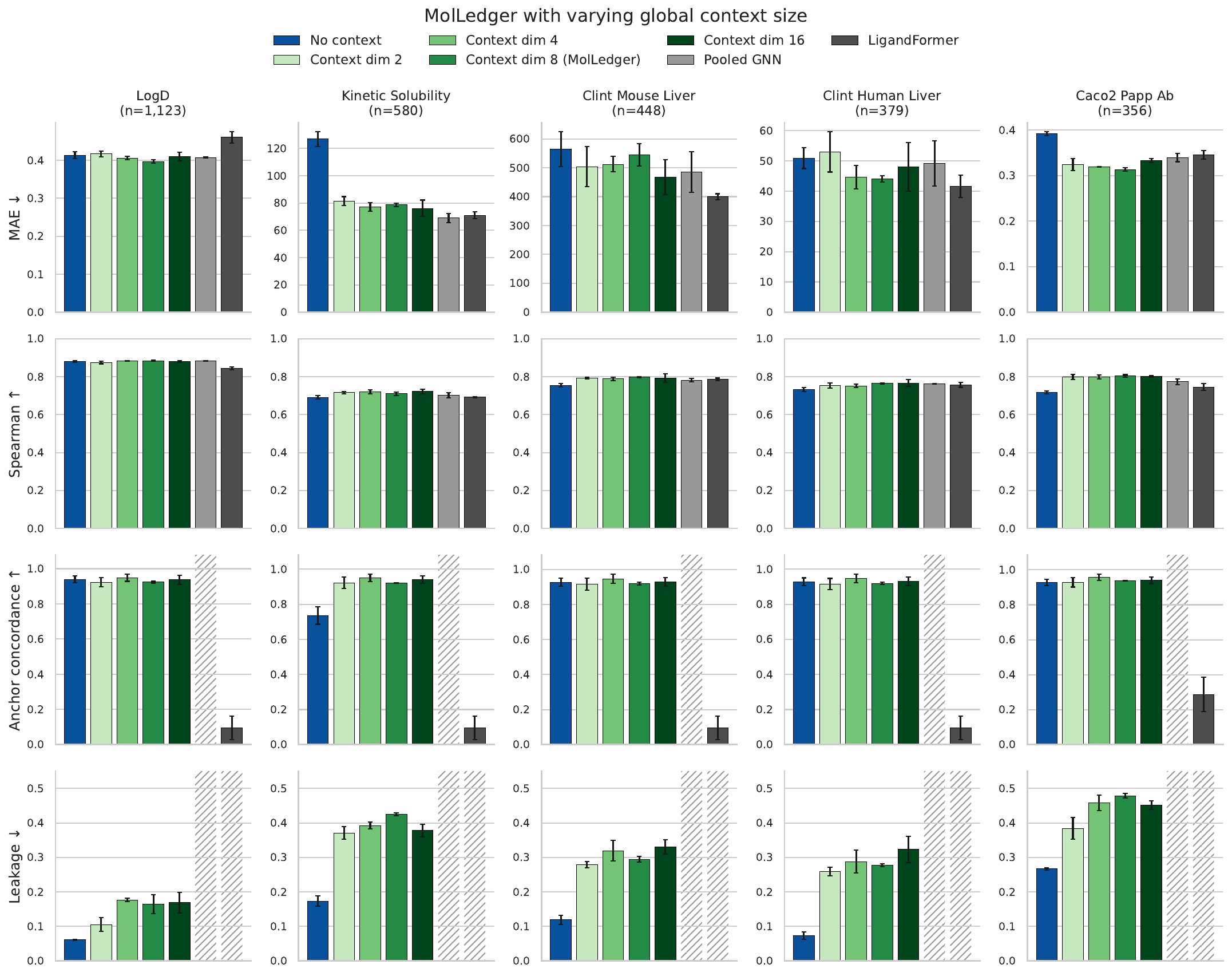}
	\caption{Performance of MolLedger when varying global context vector dimension. The no-context MolLedger uses the sum and mean form in Figure~\ref{fig:anchor_strength}. All variants of MolLedger are trained with the anchor loss. Anchor concordance and leakage are not measured for the pooled GNN as they depend on the interpretability method applied. Leakage is not defined for LigandFormer.}
	\label{fig:gctx_dim}
\end{figure}

As we introduced in Section~\ref{sec:molledger_design}, MolLedger has two novel components: the additive head with a global context vector and the loss tying each atom's attribution to a physical anchor. To evaluate the two contributions of MolLedger, we perform an in-depth analysis of how the model responds to varying the size of the global context vector and the strength of the anchor loss on validation data. Because there may be a trade-off between concordance to physical properties and predictive performance, we  examine how concordance and performance change with the strength of the anchor loss in Appendix~\ref{app:variants}. To avoid leaking information about concordance to the physical properties used for testing when tuning hyper-parameters, we evaluate concordance to the anchor rather than concordance to the held-out property for hyper-parameter tuning. Because the anchors are continuous properties, the concordance metric for each molecule is defined as the Pearson correlation between the per-atom attribution $s_i$ and the anchor $a_i$:
\begin{equation}
    \text{Anchor concordance} = \text{Pearson}\left(\left\{s_i\right\}_{i=1}^{N_{atoms}}, \left\{a_i\right\}_{i=1}^{N_{atoms}}\right)
\end{equation}
This is the term that the auxiliary loss tries to bring to 1.

Figure~\ref{fig:gctx_dim} shows the results of the analysis on global context vector size. Without a global context vector, constraining the head to be additive results in a significant increase in MAE relative to the pooled GNN and LigandFormer baselines for kinetic solubility and Caco2 A-to-B permeability. Once we add a global context vector, the MAE becomes much better on these two tasks. Concordance with the anchor also improves when the global context vector is added for kinetic solubility. As we vary the global context vector size from 2 to 16, we see that the benefit saturates around size 8, so we choose a size-8 global context vector. With this global context vector, obtaining interpretability with an additive head no longer compromises performance.

\begin{figure}
	\centering
	\includegraphics[width=\linewidth]{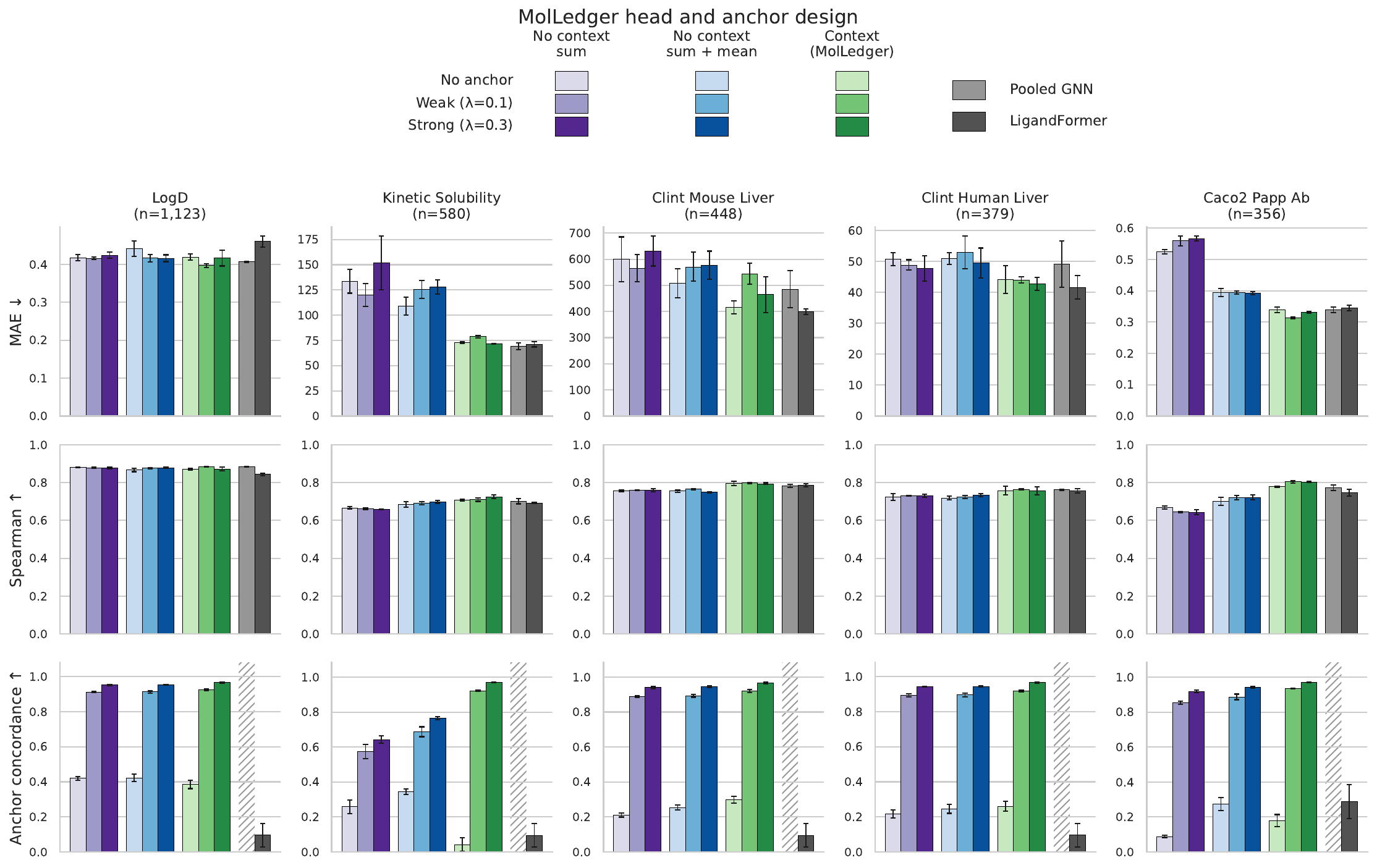}
	\caption{Performance of MolLedger with varying anchor strength. Three types of heads are evaluated for MolLedger. Anchor concordance is not measured for the pooled GNN as it depends on the interpretability method applied.}
	\label{fig:anchor_strength}
\end{figure}

Figure~\ref{fig:anchor_strength} shows how the additional constraint imposed by the anchor loss makes the per-atom attributions significantly more concordant without affecting model performance. We perform this experiment on three variants of the additive head: 1) The first variant of the additive head produces the per-atom scores with a linear function of the atom representation from message passing. The prediction is the sum of the per-atom scores $\sum_{i=1}^{N_{atoms}} W A_i$. 2) The second variant has two independent linear additive heads that both compute per-atom scores from the atom representations from message passing. The output is the sum of the sum of the per-atom scores from one head and the mean of the per-atom scores from the other head: $\sum_{i=1}^{N_{atoms}} W_1 A_i + \frac{1}{N_{atoms}} \sum_{i=1}^{N_{atoms}} W_2 A_i$. Having a sum and a mean term allows the model to adjust for molecule size . The per-atom score used to measure anchor concordance is then $W_1 A_i + \frac{1}{N} W_2 A_i$. 3) The third variant is the additive head in MolLedger with a size-8 global context vector: $\sum_{i=1}^{N_{atoms}} \phi_3\left(A_i, c\right)$, where $c$ is the global context vector. $c$ is computed from a concatenation of the sum and the mean of the atom representations from message-passing. Because $\phi_3$ is a MLP, two separate additive heads for the final output are not needed. The switch from sum to sum and mean is helpful for kinetic solubility and Caco2 A-to-B permeability as these properties are particularly affected by molecule size. The addition of a global context vector leads to further performance improvement. The global context vector can learn molecule size, as well as other informative features. For all three of these variants, the addition of even a weak anchor loss leads to a significant increase in anchor concordance, far outperforming the LigandFormer attention baseline. Increasing anchor loss strength leads to a small increase in anchor concordance with little effect on performance. When we evaluate MolLedger, we choose the best anchor strength for each seed based on the average normalized validation MAE across all tasks.

\section{Interpretability Baseline Methods}

\label{app:interp_baselines}

\textit{LigandFormer} This attention-based graph neural network was designed to interpret the contribution of each atom to an ADME prediction by taking the average of the amount of attention paid to each atom across query atoms, attention heads, and attention blocks \citep{guo2022ligandformer}. The attention mechanism only appears in the trunk, so the attention weights are the same for all tasks. This does not pose an issue for \citet{guo2022ligandformer} as they train separate models for each task. In the multi-task setting we evaluate, the attention weights that come from the trunk are the same for all tasks, so the interpretations from LigandFormer reflect which atoms are informative in general rather than which atoms are informative for a particular task. In addition, we evaluate regression tasks, whereas \citet{guo2022ligandformer} evaluate classification asks. The attention weights are designed to sum to 1 across atoms in each layer rather than sum to the prediction, so we cannot evaluate exactness for LigandFormer. A substituent may be important in one molecule of the matched pair but not in the other molecule, so the core atoms may receive different attention weights simply due to reallocation of the attention weights. This change in attention to the core cannot be called leakage. Because atoms that are predictive of high or low values may carry high attention weights, we can evaluate concordance to the magnitude of a continuous physical property. However, we cannot expect attention to be higher for atoms with one binary label versus another, so we cannot evaluate concordance to the 4 binary properties shown in Figure~\ref{fig:noncircular_concordance}. The only interpretability metric we evaluate for LigandFormer is concordance in terms of correlation between the magnitude of the continuous anchor and the attention score. This is shown in Figure~\ref{fig:gctx_dim} and~\ref{fig:anchor_strength}. LigandFormer is omitted from the interpretability plots elsewhere. 

The remaining interpretability methods are posthoc approaches that we apply to the pooled GNN. Adding molecular descriptors changes the interpretation for atom-level features from the contribution of the atom to the correction after predicting from molecular descriptors, so we choose to interpret the pooled GNN without molecular descriptors.

\begin{figure}
	\centering
	\includegraphics[width=.4\linewidth]{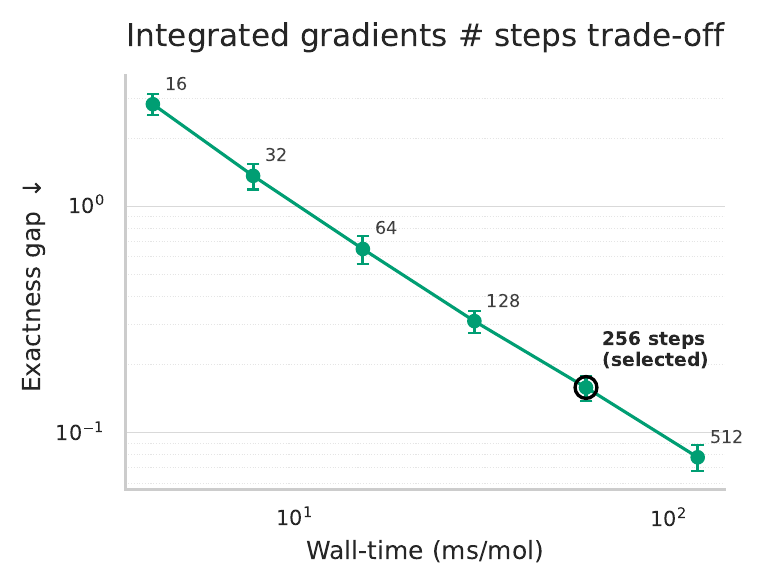}
	\caption{Exactness vs time trade-off for integrated gradients.}
	\label{fig:ig_steps}
\end{figure}

\begin{figure}
	\centering
	\includegraphics[width=.6\linewidth]{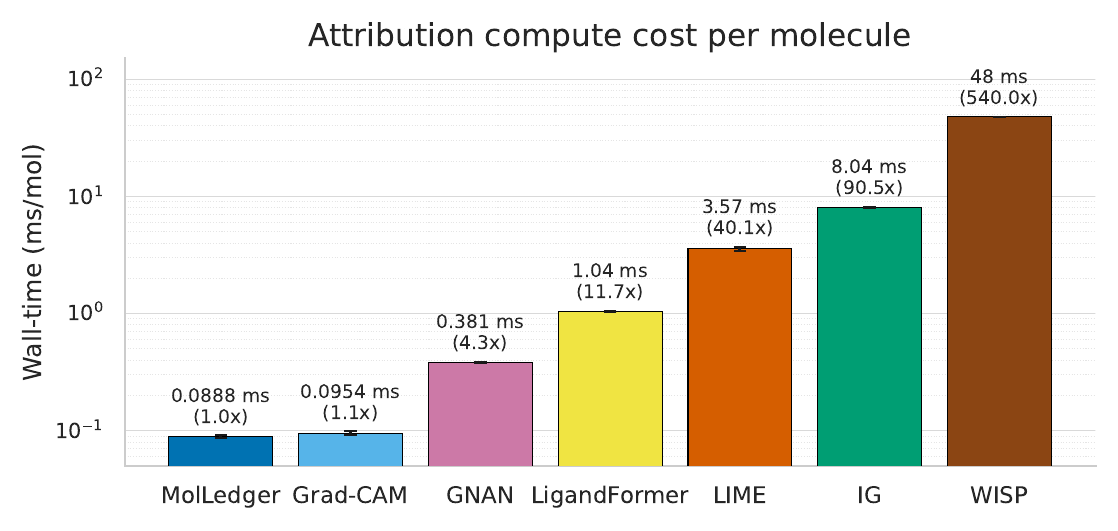}
	\caption{Time to run each interpretability method.}
	\label{fig:timing}
\end{figure}

\textit{Integrated gradients (IG)} The contribution of a feature at a certain point along a function can be approximated by the coefficients in a series of linear approximations from the origin to that point \citep{sundararajan2017axiomatic}. This idea can be extended to graph neural networks. Following \citet{ito2025improving}, we compute integrated gradients along a path from a graph with the same structure as an all-zero atom embedding to the molecule graph. The per-atom attribution is the sum of the gradients with respect to the embeddings along this path. Because the IG path starts at the all-zero baseline, we account for the prediction for that baseline molecule $\hat{y}_{base}$ when computing the exactness gap:
\begin{equation}
    \text{Exactness gap for integrated gradients} = \left| \left(\hat{y} - \hat{y}_{base}\right) -\sum_{i=1}^{N_{atoms}} s_i\right|
    \label{eq:gap_with_base}
\end{equation}
Because most matched pairs have different topologies, they also have different baseline molecules. As the difference between the predicted values for the two baseline molecules cannot be attributed to the core or the substituent, it does not appear in the definition of leakage in Equation~\ref{eq:leakage_def}. Varying the number of steps along the integrated gradients path trades off exactness for runtime. Figure~\ref{fig:ig_steps} shows how integrated gradients becomes more exact as the number of steps used to approximate the function increases. However, the time to compute integrated gradients also scales with the number of steps. We chose to run integrated gradients with 256 steps to give it more exact interpretations. This results in integrated gradients taking more than twice as long compared to other interpretability methods besides WISP as shown in Figure~\ref{fig:timing}.

\textit{Grad-CAM} While IG follows a linear approximation in the embedding space that is inputted to a model, gradient-weighted class activation mapping (Grad-CAM) creates a linear approximation in the activation space outputted before the prediction head \citep{selvaraju2017grad}. Originally designed for classification tasks with convolutional neural networks, Grad-CAM computes the gradient of the logit for a class with respect to the activation in a feature channel averaged across pixels in the activation space. Then, to compute how much a pixel contributes to the prediction of a class across all feature channels, Grad-CAM sums the activations across channels weighted by these gradients and gates this sum with ReLU. \citet{wei2022interpretable} adapt Grad-CAM to ADME tasks by taking the gradient of the prediction with respect to the output from the final message-passing layer in the GNN. For task $c$, the weight on activation dimension $k$ in this GNN embedding is the average gradient across atoms:
\begin{equation}
    \alpha^c_k = \frac{1}{N_{atoms}} \sum_{i=1}^{N_{atoms}} \frac{\partial y^c}{\partial A_i^k}
\end{equation}
The contribution of atom $i$ to task $c$ is then $s_i = \sum_k \alpha^c_k A_i^k$. For regression tasks, the contribution of an atom can be positive or negative, so the ReLU is dropped. Similar to IG, the exactness gap measures the difference between the sum of these per-atom attributions and the prediction from a baseline molecule. For Grad-CAM, this baseline molecule has all-zero activations from message passing rather than an all-zero input embedding. When we examine the definition for Grad-CAM, we can observe the two terms that lead to leakage for a core atom with activation $\mathbf{A}_i^{\left(a\right)}$ in molecule $a$ and $\mathbf{A}_i^{\left(b\right)}$ in molecule $b$:
\begin{align}
    \text{Leakage for core atom i in Grad-CAM} &= \left| \mathbf{\alpha}^{\left(a\right)} A_i^{\left(a\right)} - \mathbf{\alpha}^{\left(b\right)} A_i^{\left(b\right)} \right| \\
    &= \left| \mathbf{\alpha}^{\left(a\right)} \left(A_i^{\left(a\right)} - A_i^{\left(b\right)}\right) + \left(\mathbf{\alpha}^{\left(a\right)} - \mathbf{\alpha}^{\left(b\right)}\right) A_i^{\left(b\right)} \right|
\end{align}
The first term reflects the change in activations across the two molecules, and the second term reflects the change in gradients with respect to the activation across the pair.

\textit{LIME} Local interpretable model-agnostic explanation (LIME) samples perturbations of the input, computes model predictions for these perturbed inputs, and fits a linear model on these perturbed input-output pairs \citep{ribeiro2016should}. \citet{jamrozik2024admet} apply LIME to interpret the contributions of substructure fingerprints to ADME predictions from graph neural networks. Because we are interested in per-atom contributions rather than per-substructure contributions, we perform our perturbations to $\tilde{\mathbf{Z}} \in \left\{0, 1\right\}^{N_{atoms}}$, a binary mask indicating whether each atom is present or absent.  We gather 300 perturbed samples in this binary space and fit the linear model $\tilde{\mathbf{Y}} = \hat{\mathbf{W}} \tilde{\mathbf{Z}} + \hat{b}$, where  $\hat{\mathbf{W}}$ is the per-atom contribution. The bias term $\hat{b}$ is treated like $\hat{y}_{base}$ in Equation~\ref{eq:gap_with_base} when estimating the exactness gap for LIME. The bias is also omitted when computing leakage.

\textit{WISP} The interpretability method employed in the workflow for interpretability scoring using matched molecular pairs (WISP) introduced in \citet{janssen2026machine} and \citet{janssen2026machine2} is a form of occlusion adapted specifically for molecules. To compute the attribution for an individual atom, the atom is swapped for H, B, C, N, O, F, Si, P, S, Cl, Br, or I where the substitution is valid with implicit hydrogens. The average change in prediction across the valid substitutions is the attribution to that atom. Because there is no single baseline molecule, the exactness gap directly compares the sum of these per-atom attributions to the prediction. The long runtime for WISP in Figure~\ref{fig:timing} comes from constructing the modified molecules in RDKit and checking for validity.

\section{Additional Case Studies}

\label{app:case_studies}

\begin{figure}
	\centering
	\includegraphics[width=.85\linewidth]{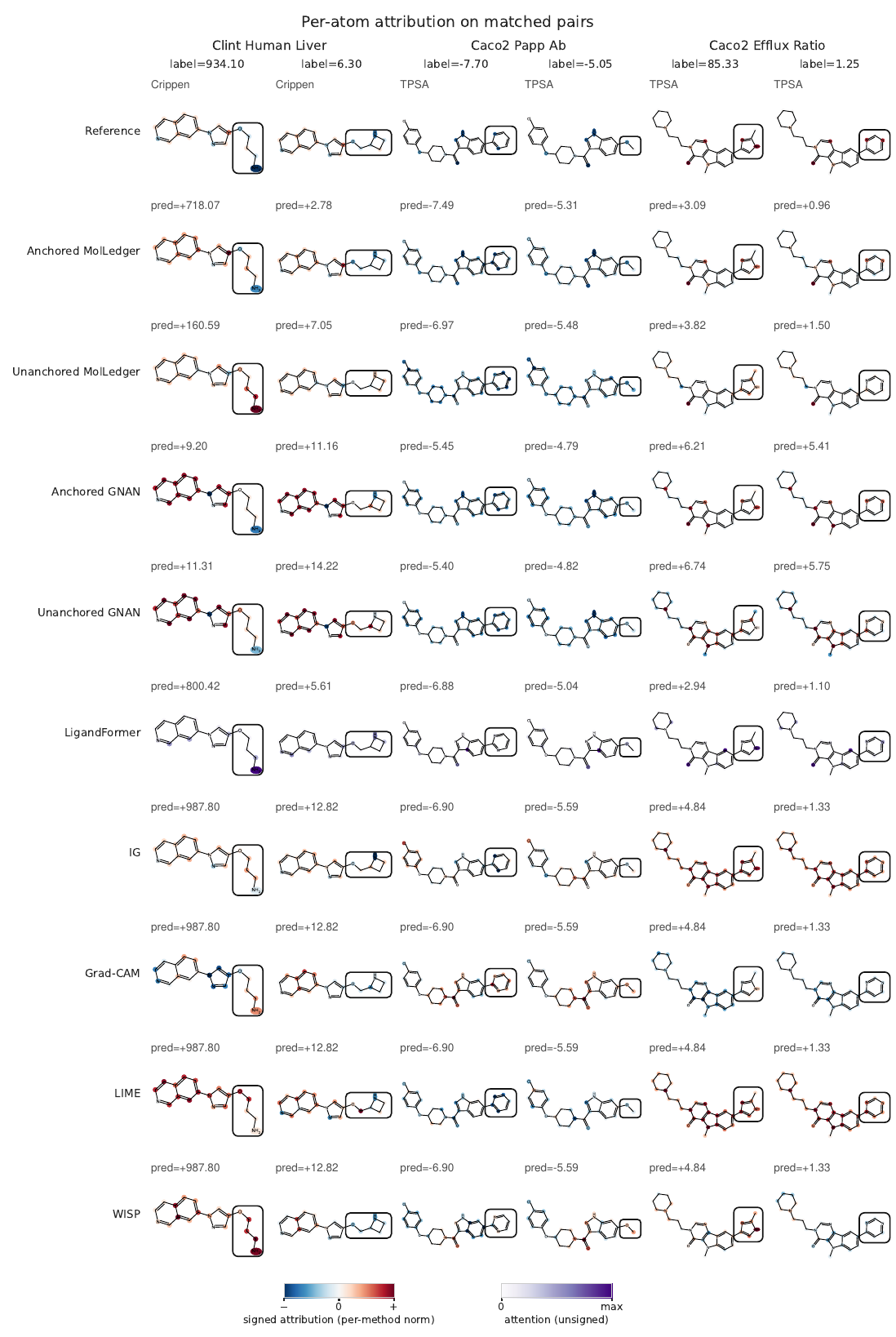}
	\caption{Per-atom attributions on matched molecular pairs with the largest property change for intrinsic clearance in human liver cells, Caco2 A-to-B permeability assay, and Caco2 efflux ratio. Swapped fragments are shown in a box.}
	\label{fig:visual_other}
\end{figure}

\begin{figure}
	\centering
	\includegraphics[width=.85\linewidth]{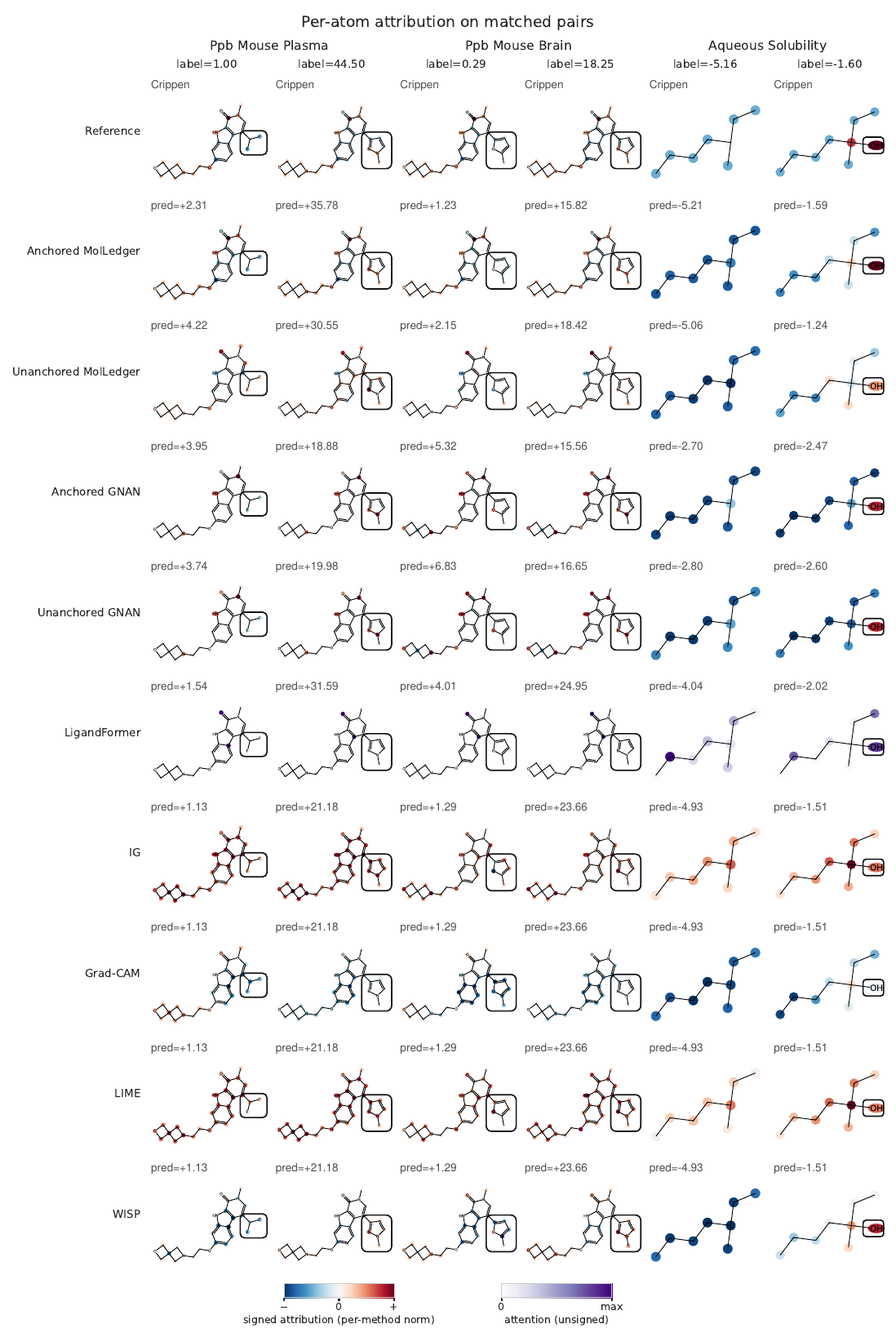}
	\caption{Per-atom attributions on matched molecular pairs with the largest property change for plasma protein binding in mouse plasma and brain. Swapped fragments are shown in a box. We select a special case for aqueous solubility where a new subgroup is introduced in place of a hydrogen, so the left molecule does not have a box.}
	\label{fig:visual_ppb}
\end{figure}

In addition to the three case studies in Section~\ref{sec:case_studies}, we analyze 6 examples for other tasks shown in Figures~\ref{fig:visual_other} and~\ref{fig:visual_ppb}. Starting with the leftmost example in Figure~\ref{fig:visual_other}, this matched molecular pair for intrinsic clearance in human liver cells is a more nuanced case where the anchor is misleading. The aminopropyl chain in the molecule on the left has negative Crippen scores because it is lipophobic. However, the flexible primary amine chain means Cyp450 could perform O-dealkylation or oxidative deamination and break the chain. Closing a ring to block metabolism is a common med-chem strategy for reducing clearance. The rigid azetidine on the right is much less accessible for oxidation. Unanchored MolLedger is able to learn about the flexible primary amine structure and assign positive attributions to all the atoms in the aminopropyl chain, but it has not learned that the azetidine should have negative scores. On the other hand, because the anchors influence the per-atom attributions, anchored MolLedger is able to attribute negative scores to all the atoms in the azetidine ring, but it places negative attributions on the polar oxygen and amino group as well. In contrast, for GNAN, even the unanchored model attributes a negative score to the amino group despite receiving information about the amino group being at the end of a chain from GNAN's topology weighting. GNAN also does not place negative weights around the azetidine besides at the nitrogen. LigandFormer attends strongly to the terminal amino group, but the attention weight does not indicate whether the model believes the amino group leads to faster or slower clearance. Grad-CAM and WISP on the pooled model place positive contributions on all atoms in the aminopropyl chain and negative contributions on all atoms in the azetidine, making them strong interpretations in this case. LIME interprets the contribution of the aminopropyl correctly but not the azetidine, and IG is incorrect on both molecules. These different interpretations of the same model show that we cannot trust the interpretations when they are not directly tied to the prediction.

In the example for Caco2 A-to-B permeability, the swap changes a pyrimidine to a methoxy. The pyrimidine has 2 hydrogen bond acceptors (HBA) that form a Seelig motif and bind to the P-gp transport protein responsible for actively pumping drugs out of a cell \citep{seelig1998general}. In contrast, the ether is much less polar, and the Seelig motif is no longer present in the compound on the right. This means the drug tends to stay inside the cell much more. Aligning to the TPSA anchor is well-motivated in this case. Anchored MolLedger places more of the negative attribution on the two HBAs, whereas the unanchored model is more diffuse around the ring. Similar trends are observed for GNAN. IG and LIME also seem to capture the right attributions, whereas Grad-CAM and WISP report the wrong direction and LigandFormer does not attend to the two HBAs.

The next example for Caco2 efflux ratio swaps a methylimidazole to a pyrimidine. The $sp^2$-hybridized nitrogen in the imidazole is a strong HBA that anchors the molecule to the P-gp binding site. Once the molecule is anchored to the binding site, a second HBA in the core of the molecule can then form the other part of the Seelig motif. In contrast, the pyrimidine in the molecule on the right cannot act as an anchor in the binding site. The pyrimidine is attached to the core at the carbon next to only one of the nitrogens rather than at the carbon between the two nitrogens as in the example for Caco2 A-to-B. The core displaces the lone pair on the nearby nitrogen, and the other nitrogen is not strong enough to act as the HBA that anchors the molecule in the P-gp binding site. In this case, the TPSA anchors are only partially informative, as they are not strong indicators of donor strength. Anchored MolLedger and anchored GNAN follow the TPSA anchors and attach significance to both nitrogens in both fragments, rather than only to the HBA nitrogen in the imidazole. LigandFormer, Grad-CAM, and WISP put weight on the wrong nitrogen in the imidazole, while IG and LIME place weights all around the rings. All models cannot predict the large Caco2 efflux ratio for the molecule on the left. Even when the pooled GNN does not have an additive constraint, it is still unable to predict the exponential increase that occurs with the methylimidazole compared to the pyrimidine. This is an example of an activity cliff, which is challenging to predict \citep{stumpfe2014recent}. Part of the challenge here is the introduction of a HBA in the methylimidazole affects the value of HBAs in the core. In this case, leakage into the core is justified, but the interpretability methods reflect that the models do not treat the core differently between these two molecules.

In Figure~\ref{fig:visual_ppb}, both examples for plasma protein binding in mouse plasma and brain show swaps to methylprazole, a group with two nitrogens that act as HBAs. The left molecule in the mouse plasma pair is fairly lipophilic as the two fluorines are poor HBAs due to electronegativity holding the lone pairs tightly. The left molecule in the mouse brain pair is also lipophilic as the oxygen is only weakly polar in the ring. Both of these swaps introduce much stronger HBAs that make the second compound less lipophilic and increase the fraction unbound. Anchored MolLedger, both versions of GNAN, and WASP attribute these changes to the responsible atoms. In contrast, unanchored MolLedger suggests that the two fluorines increase the fraction unbound as it misses how fluorines are weak HBAs in this structure. LigandFormer does not attend to the substituents. Besides a negative attribution placed on the polar oxygen, IG and LIME place positive attributions throughout both substituents, and Grad-CAM places negative attributions throughout both. This once again reflects the issue with posthoc interpretability methods when they yield conflicting interpretations of the same model.

Finally, for aqueous solubility, we select an example of a special case we defined beyond rdkit's matched pair analysis. This pair adds a polar hydroxyl in place of a hydrogen. The left molecule has no heavy atom in its substituent group, and thus the change can only be interpreted with the weight on the hydroxyl in the new molecule. All methods attribute the increase in aqueous solubility to either the hydroxyl or the hydroxyl-bearing carbon that is affected by the swap.

\section{Additional Results}

\label{app:more_results}

\begin{figure}[h]
	\centering
	\includegraphics[width=\linewidth]{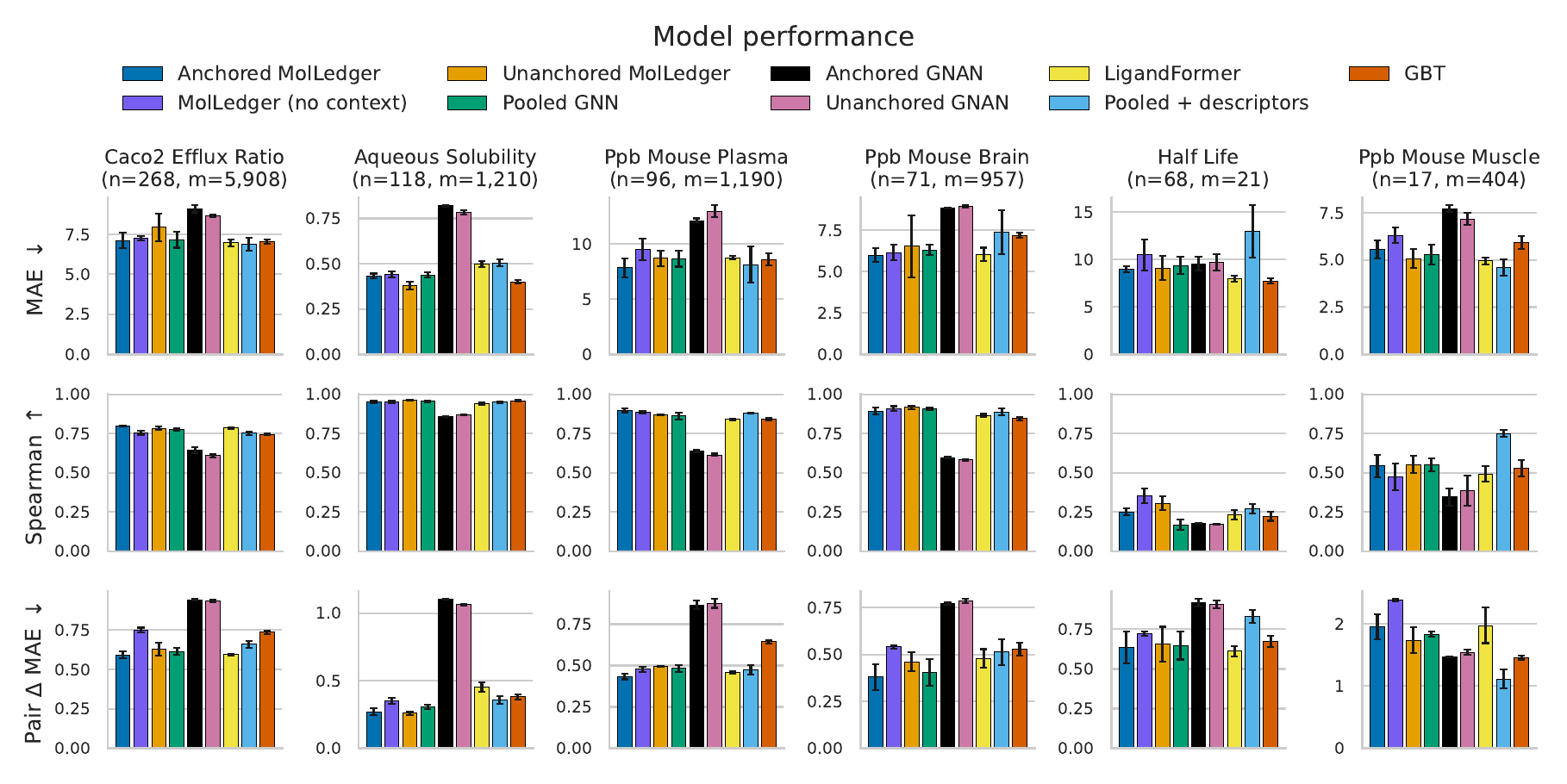}
	\caption{Performance on smaller tasks as a continuation of Figure~\ref{fig:performance}. $n$ is the number of test samples, and $m$ is the number of matched pairs with a molecule from the test set. Mean and standard deviation across 3 random seeds.}
	\label{fig:performance_other}
\end{figure}

\begin{figure}[h]
	\centering
	\includegraphics[width=\linewidth]{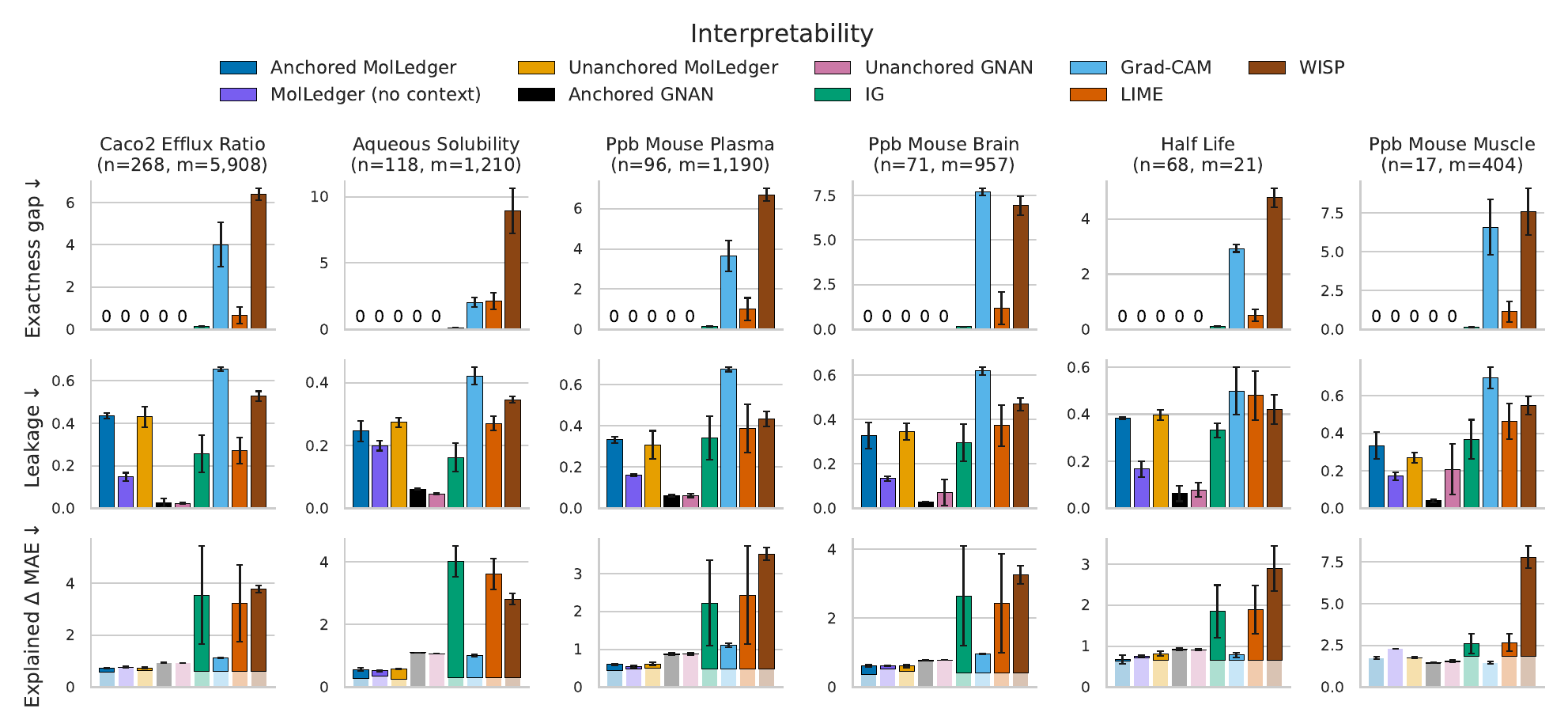}
	\caption{Interpretability metrics on smaller tasks as a continuation of Figure~\ref{fig:interpretability}. $n$ is the number of test samples, and $m$ is the number of matched pairs with a molecule from the test set. The transparent part of the explained $\Delta$ MAE bars is the MAE for the predicted differences shown in Figure~\ref{fig:performance_other}. The solid part represents attribution error. Mean and standard deviation across 3 random seeds.}
	\label{fig:interpretability_other}
\end{figure}

\begin{table}[h]
\centering
\small
\caption{MAE. These are the numbers in the top rows of Figures~\ref{fig:performance} and~\ref{fig:performance_other}. Best shown in bold.}
\label{tab:perf-mae}
\vspace{0.2cm}
\begin{tabular}{lcccccc}
\toprule
Model & LogD & Kin.\ sol. & CL$_{int}$ mouse & CL$_{int}$ human & Caco2 P$_{app}$ & Caco2 efflux \\
\midrule
MolLedger & \textbf{0.40 (0.00)} & 76.0 (3.3) & 533.0 (63.4) & 60.3 (7.5) & 0.35 (0.00) & 7.12 (0.46) \\
MolLedg. no ctx & 0.41 (0.01) & 146.5 (10.2) & 520.6 (46.3) & 60.0 (4.0) & 0.44 (0.02) & 7.25 (0.13) \\
Unanch MolLedg. & 0.42 (0.01) & 78.6 (1.5) & 411.6 (25.4) & 51.4 (3.5) & \textbf{0.34 (0.01)} & 7.94 (0.86) \\
Pooled GNN & \textbf{0.40 (0.00)} & \textbf{69.2 (2.3)} & 460.3 (37.5) & 70.8 (16.1) & 0.35 (0.01) & 7.16 (0.50) \\
Anch GNAN & 0.80 (0.00) & 111.5 (1.6) & 435.8 (2.8) & 55.5 (0.2) & 0.63 (0.00) & 9.10 (0.23) \\
Unanch GNAN & 0.80 (0.01) & 111.6 (1.6) & 428.3 (3.2) & 54.7 (0.4) & 0.62 (0.00) & 8.64 (0.07) \\
LigandFormer & 0.45 (0.01) & 71.0 (1.4) & 372.4 (8.9) & \textbf{45.8 (2.5)} & 0.36 (0.01) & 6.97 (0.21) \\
Pooled + desc & 0.50 (0.01) & 77.8 (4.6) & 418.2 (38.9) & 56.1 (9.5) & 0.38 (0.01) & \textbf{6.88 (0.41)} \\
GBT & 0.55 (0.00) & 73.8 (0.5) & \textbf{358.0 (0.4)} & 49.6 (0.8) & 0.39 (0.01) & 7.03 (0.12) \\
\bottomrule
\end{tabular}
\\[4pt]
\begin{tabular}{lccccc}
\toprule
Model & Aq.\ sol. & PPB plasma & PPB brain & Half-life & PPB muscle \\
\midrule
MolLedger & 0.43 (0.01) & \textbf{7.86 (0.87)} & \textbf{5.99 (0.43)} & 8.99 (0.31) & 5.55 (0.50) \\
MolLedg. no ctx & 0.44 (0.02) & 9.54 (0.96) & 6.16 (0.49) & 10.49 (1.61) & 6.31 (0.41) \\
Unanch MolLedg. & \textbf{0.38 (0.02)} & 8.71 (0.70) & 6.53 (1.86) & 9.08 (1.28) & 5.07 (0.48) \\
Pooled GNN & 0.44 (0.02) & 8.69 (0.73) & 6.31 (0.33) & 9.37 (0.90) & 5.28 (0.54) \\
Anch GNAN & 0.82 (0.01) & 12.15 (0.19) & 8.84 (0.02) & 9.56 (0.69) & 7.73 (0.19) \\
Unanch GNAN & 0.78 (0.01) & 13.01 (0.56) & 8.95 (0.09) & 9.69 (0.84) & 7.17 (0.30) \\
LigandFormer & 0.50 (0.02) & 8.77 (0.14) & 6.04 (0.40) & 7.97 (0.32) & 4.95 (0.18) \\
Pooled + desc & 0.51 (0.02) & 8.15 (1.62) & 7.40 (1.33) & 12.98 (2.78) & \textbf{4.59 (0.44)} \\
GBT & 0.40 (0.01) & 8.61 (0.55) & 7.20 (0.14) & \textbf{7.73 (0.29)} & 5.92 (0.34) \\
\bottomrule
\end{tabular}
\end{table}

\begin{table}[h]
\centering
\small
\caption{Spearman. These are the numbers in the middle rows of Figures~\ref{fig:performance} and~\ref{fig:performance_other}. Best shown in bold.}
\label{tab:perf-spearman}
\vspace{0.2cm}
\begin{tabular}{lcccccc}
\toprule
Model & LogD & Kin.\ sol. & CL$_{int}$ mouse & CL$_{int}$ human & Caco2 P$_{app}$ & Caco2 efflux \\
\midrule
MolLedger & \textbf{0.89 (0.00)} & \textbf{0.72 (0.01)} & \textbf{0.77 (0.02)} & \textbf{0.79 (0.02)} & 0.79 (0.01) & \textbf{0.80 (0.00)} \\
MolLedg. no ctx & \textbf{0.88 (0.01)} & 0.69 (0.01) & 0.76 (0.01) & 0.75 (0.00) & 0.67 (0.01) & 0.75 (0.01) \\
Unanch MolLedg. & 0.87 (0.01) & 0.71 (0.01) & 0.77 (0.02) & 0.78 (0.01) & \textbf{0.80 (0.01)} & 0.78 (0.01) \\
Pooled GNN & \textbf{0.89 (0.00)} & 0.71 (0.00) & \textbf{0.78 (0.01)} & 0.78 (0.01) & 0.77 (0.00) & 0.78 (0.01) \\
Anch GNAN & 0.58 (0.00) & 0.39 (0.03) & 0.53 (0.00) & 0.56 (0.00) & 0.45 (0.00) & 0.64 (0.02) \\
Unanch GNAN & 0.58 (0.00) & 0.39 (0.02) & 0.53 (0.00) & 0.56 (0.01) & 0.45 (0.01) & 0.61 (0.01) \\
LigandFormer & 0.85 (0.00) & 0.68 (0.01) & 0.76 (0.02) & 0.78 (0.01) & 0.75 (0.01) & 0.79 (0.01) \\
Pooled + desc & 0.83 (0.01) & 0.67 (0.01) & 0.74 (0.01) & 0.72 (0.03) & 0.75 (0.01) & 0.75 (0.01) \\
GBT & 0.79 (0.00) & 0.62 (0.01) & 0.73 (0.01) & 0.71 (0.02) & 0.72 (0.01) & 0.75 (0.01) \\
\bottomrule
\end{tabular}
\\[4pt]
\begin{tabular}{lccccc}
\toprule
Model & Aq.\ sol. & PPB plasma & PPB brain & Half-life & PPB muscle \\
\midrule
MolLedger & 0.95 (0.01) & \textbf{0.90 (0.01)} & 0.89 (0.02) & 0.25 (0.02) & 0.54 (0.07) \\
MolLedg. no ctx & 0.95 (0.01) & 0.89 (0.01) & 0.91 (0.01) & \textbf{0.35 (0.05)} & 0.47 (0.08) \\
Unanch MolLedg. & \textbf{0.96 (0.00)} & 0.87 (0.00) & \textbf{0.92 (0.01)} & 0.31 (0.04) & 0.55 (0.05) \\
Pooled GNN & 0.96 (0.00) & 0.86 (0.02) & 0.91 (0.01) & 0.17 (0.03) & 0.55 (0.04) \\
Anch GNAN & 0.86 (0.00) & 0.64 (0.01) & 0.60 (0.01) & 0.18 (0.00) & 0.35 (0.05) \\
Unanch GNAN & 0.87 (0.00) & 0.61 (0.01) & 0.58 (0.01) & 0.17 (0.00) & 0.39 (0.10) \\
LigandFormer & 0.94 (0.01) & 0.84 (0.01) & 0.86 (0.01) & 0.23 (0.03) & 0.49 (0.05) \\
Pooled + desc & 0.95 (0.00) & 0.88 (0.00) & 0.89 (0.02) & 0.27 (0.03) & \textbf{0.75 (0.02)} \\
GBT & \textbf{0.96 (0.00)} & 0.84 (0.01) & 0.85 (0.01) & 0.22 (0.03) & 0.53 (0.05) \\
\bottomrule
\end{tabular}
\end{table}

\begin{table}
\centering
\small
\caption{Matched-pair $\Delta$ MAE. These are the numbers in the bottom rows of Figures~\ref{fig:performance} and~\ref{fig:performance_other}. Best shown in bold.}
\vspace{0.2cm}
\label{tab:perf-matchedpair}
\begin{tabular}{lcccccc}
\toprule
Model & LogD & Kin.\ sol. & CL$_{int}$ mouse & CL$_{int}$ human & Caco2 P$_{app}$ & Caco2 efflux \\
\midrule
MolLedger & \textbf{0.35 (0.01)} & 0.74 (0.02) & 0.95 (0.02) & 0.78 (0.04) & \textbf{0.32 (0.00)} & \textbf{0.59 (0.02)} \\
MolLedg. no ctx & 0.37 (0.01) & 0.96 (0.05) & 1.01 (0.01) & 0.91 (0.02) & 0.42 (0.01) & 0.75 (0.01) \\
Unanch MolLedg. & 0.38 (0.02) & 0.81 (0.01) & 0.99 (0.02) & 0.84 (0.02) & 0.33 (0.03) & 0.63 (0.04) \\
Pooled GNN & 0.36 (0.00) & 0.72 (0.02) & 0.95 (0.04) & 0.81 (0.03) & 0.34 (0.01) & 0.61 (0.02) \\
Anch GNAN & 0.66 (0.00) & 1.05 (0.02) & 1.26 (0.00) & 1.09 (0.01) & 0.55 (0.00) & 0.94 (0.01) \\
Unanch GNAN & 0.66 (0.00) & 1.05 (0.02) & 1.26 (0.00) & 1.09 (0.01) & 0.54 (0.01) & 0.93 (0.01) \\
LigandFormer & 0.36 (0.00) & \textbf{0.67 (0.01)} & \textbf{0.84 (0.02)} & \textbf{0.72 (0.04)} & 0.32 (0.01) & \textbf{0.59 (0.01)} \\
Pooled + desc & 0.42 (0.01) & 0.77 (0.02) & 0.97 (0.06) & 0.88 (0.01) & 0.39 (0.00) & 0.66 (0.02) \\
GBT & 0.45 (0.01) & 0.79 (0.01) & 0.95 (0.01) & 0.88 (0.01) & 0.40 (0.01) & 0.74 (0.01) \\
\bottomrule
\end{tabular}
\\[4pt]
\begin{tabular}{lccccc}
\toprule
Model & Aq.\ sol. & PPB plasma & PPB brain & Half-life & PPB muscle \\
\midrule
MolLedger & 0.27 (0.02) & \textbf{0.43 (0.02)} & \textbf{0.38 (0.07)} & 0.64 (0.10) & 1.95 (0.20) \\
MolLedg. no ctx & 0.35 (0.02) & 0.48 (0.02) & 0.54 (0.01) & 0.72 (0.01) & 2.38 (0.02) \\
Unanch MolLedg. & \textbf{0.26 (0.01)} & 0.50 (0.00) & 0.46 (0.05) & 0.66 (0.11) & 1.74 (0.20) \\
Pooled GNN & 0.31 (0.02) & 0.48 (0.02) & 0.41 (0.07) & 0.65 (0.09) & 1.83 (0.04) \\
Anch GNAN & 1.10 (0.00) & 0.87 (0.03) & 0.77 (0.01) & 0.92 (0.02) & 1.46 (0.01) \\
Unanch GNAN & 1.06 (0.01) & 0.88 (0.03) & 0.79 (0.01) & 0.91 (0.02) & 1.54 (0.04) \\
LigandFormer & 0.45 (0.04) & 0.46 (0.01) & 0.48 (0.05) & \textbf{0.61 (0.03)} & 1.97 (0.28) \\
Pooled + desc & 0.36 (0.03) & 0.47 (0.03) & 0.51 (0.07) & 0.83 (0.04) & \textbf{1.11 (0.15)} \\
GBT & 0.38 (0.02) & 0.64 (0.01) & 0.53 (0.03) & 0.67 (0.03) & 1.45 (0.03) \\
\bottomrule
\end{tabular}
\end{table}

\begin{table}
\centering
\small
\caption{Concordance to physical properties not used during training. These are the numbers plotted in Figure~\ref{fig:noncircular_concordance} with bold being best.}
\vspace{0.2cm}
\label{tab:noncircular-concordance}
\begin{tabular}{lcccc}
\toprule
Method & LogD (hydrophobic) & PPB plasma (aromatic) & Caco2 P$_{app}$ (HBD) & Caco2 efflux (HBA) \\
\midrule
MolLedger & 0.86 (0.01) & \textbf{0.76 (0.00)} & \textbf{0.94 (0.00)} & 0.94 (0.00) \\
MolLedg. no ctx & 0.86 (0.00) & \textbf{0.76 (0.00)} & \textbf{0.94 (0.00)} & \textbf{0.96 (0.00)} \\
Unanch MolLedg. & 0.73 (0.02) & 0.63 (0.04) & 0.66 (0.05) & 0.64 (0.02) \\
Anch GNAN & \textbf{0.97 (0.00)} & 0.72 (0.02) & 0.84 (0.01) & \textbf{0.95 (0.01)} \\
Unanch GNAN & 0.95 (0.00) & 0.64 (0.03) & 0.76 (0.02) & 0.65 (0.01) \\
LigandFormer & 0.30 (0.05) & 0.52 (0.08) & 0.84 (0.04) & 0.68 (0.07) \\
IG & 0.69 (0.10) & 0.48 (0.08) & 0.59 (0.09) & 0.63 (0.04) \\
Grad-CAM & 0.72 (0.01) & 0.68 (0.03) & 0.54 (0.01) & 0.62 (0.02) \\
LIME & 0.55 (0.08) & 0.46 (0.10) & 0.60 (0.07) & 0.58 (0.06) \\
WISP & 0.80 (0.00) & 0.70 (0.03) & 0.64 (0.03) & 0.64 (0.01) \\
\bottomrule
\end{tabular}
\end{table}

\begin{table}
\centering
\small
\caption{Exactness gap. These are the numbers in the middle rows of Figures~\ref{fig:interpretability} and~\ref{fig:interpretability_other}. Best shown in bold.}
\label{tab:interp-exactness}
\vspace{0.2cm}
\begin{tabular}{lcccccc}
\toprule
Method & LogD & Kin.\ sol. & CL$_{int}$ mouse & CL$_{int}$ human & Caco2 P$_{app}$ & Caco2 efflux \\
\midrule
MolLedger & \textbf{0.00 (0.00)} & \textbf{0.00 (0.00)} & \textbf{0.00 (0.00)} & \textbf{0.00 (0.00)} & \textbf{0.00 (0.00)} & \textbf{0.00 (0.00)} \\
MolLedg. no ctx & \textbf{0.00 (0.00)} & \textbf{0.00 (0.00)} & \textbf{0.00 (0.00)} & \textbf{0.00 (0.00)} & \textbf{0.00 (0.00)} & \textbf{0.00 (0.00)} \\
Unanch MolLedg. & \textbf{0.00 (0.00)} & \textbf{0.00 (0.00)} & \textbf{0.00 (0.00)} & \textbf{0.00 (0.00)} & \textbf{0.00 (0.00)} & \textbf{0.00 (0.00)} \\
Anch GNAN & \textbf{0.00 (0.00)} & \textbf{0.00 (0.00)} & \textbf{0.00 (0.00)} & \textbf{0.00 (0.00)} & \textbf{0.00 (0.00)} & \textbf{0.00 (0.00)} \\
Unanch GNAN & \textbf{0.00 (0.00)} & \textbf{0.00 (0.00)} & \textbf{0.00 (0.00)} & \textbf{0.00 (0.00)} & \textbf{0.00 (0.00)} & \textbf{0.00 (0.00)} \\
IG & 0.23 (0.07) & 0.57 (0.08) & 0.48 (0.07) & 0.49 (0.10) & 0.20 (0.05) & 0.33 (0.06) \\
Grad-CAM & 1.15 (0.09) & 1.04 (0.08) & 2.26 (0.17) & 2.28 (0.05) & 0.83 (0.06) & 1.40 (0.13) \\
LIME & 0.78 (0.07) & 2.13 (0.56) & 2.15 (1.36) & 1.49 (0.24) & 2.32 (1.75) & 2.28 (0.98) \\
WISP & 3.12 (0.28) & 6.47 (0.36) & 8.71 (0.23) & 7.72 (0.62) & 3.06 (0.10) & 5.40 (0.24) \\
\bottomrule
\end{tabular}
\\[4pt]
\begin{tabular}{lccccc}
\toprule
Method & Aq.\ sol. & PPB plasma & PPB brain & Half-life & PPB muscle \\
\midrule
MolLedger & \textbf{0.00 (0.00)} & \textbf{0.00 (0.00)} & \textbf{0.00 (0.00)} & \textbf{0.00 (0.00)} & \textbf{0.00 (0.00)} \\
MolLedg. no ctx & \textbf{0.00 (0.00)} & \textbf{0.00 (0.00)} & \textbf{0.00 (0.00)} & \textbf{0.00 (0.00)} & \textbf{0.00 (0.00)} \\
Unanch MolLedg. & \textbf{0.00 (0.00)} & \textbf{0.00 (0.00)} & \textbf{0.00 (0.00)} & \textbf{0.00 (0.00)} & \textbf{0.00 (0.00)} \\
Anch GNAN & \textbf{0.00 (0.00)} & \textbf{0.00 (0.00)} & \textbf{0.00 (0.00)} & \textbf{0.00 (0.00)} & \textbf{0.00 (0.00)} \\
Unanch GNAN & \textbf{0.00 (0.00)} & \textbf{0.00 (0.00)} & \textbf{0.00 (0.00)} & \textbf{0.00 (0.00)} & \textbf{0.00 (0.00)} \\
IG & 0.08 (0.01) & 0.35 (0.05) & 0.34 (0.06) & 0.23 (0.06) & 0.29 (0.04) \\
Grad-CAM & 0.97 (0.10) & 1.45 (0.09) & 1.21 (0.03) & 0.75 (0.33) & 1.61 (0.20) \\
LIME & 3.57 (0.78) & 2.06 (0.99) & 2.10 (0.97) & 1.90 (0.70) & 2.12 (0.29) \\
WISP & 3.58 (0.24) & 4.44 (0.37) & 4.44 (0.66) & 2.80 (0.70) & 8.14 (0.28) \\
\bottomrule
\end{tabular}
\end{table}

\begin{table}
\centering
\small
\caption{Leakage. These are the numbers in the top rows of Figures~\ref{fig:interpretability} and~\ref{fig:interpretability_other}. Best shown in bold.}
\label{tab:interp-leakage}
\vspace{0.2cm}
\begin{tabular}{lcccccc}
\toprule
Method & LogD & Kin.\ sol. & CL$_{int}$ mouse & CL$_{int}$ human & Caco2 P$_{app}$ & Caco2 efflux \\
\midrule
MolLedger & 0.16 (0.03) & 0.43 (0.00) & 0.29 (0.01) & 0.28 (0.00) & 0.48 (0.01) & 0.44 (0.01) \\
MolLedg. no ctx & 0.06 (0.00) & 0.18 (0.01) & \textbf{0.13 (0.00)} & 0.07 (0.01) & 0.28 (0.01) & 0.15 (0.02) \\
Unanch MolLedg. & 0.19 (0.01) & 0.43 (0.05) & 0.29 (0.03) & 0.24 (0.01) & 0.38 (0.01) & 0.43 (0.05) \\
Anch GNAN & \textbf{0.02 (0.00)} & \textbf{0.17 (0.01)} & 0.17 (0.00) & \textbf{0.05 (0.02)} & 0.32 (0.01) & 0.03 (0.02) \\
Unanch GNAN & \textbf{0.02 (0.00)} & \textbf{0.17 (0.02)} & 0.26 (0.00) & 0.12 (0.03) & 0.35 (0.01) & \textbf{0.02 (0.00)} \\
IG & 0.35 (0.02) & 0.46 (0.06) & 0.33 (0.12) & 0.45 (0.03) & \textbf{0.26 (0.16)} & 0.26 (0.09) \\
Grad-CAM & 0.64 (0.03) & 0.58 (0.02) & 0.64 (0.00) & 0.66 (0.01) & 0.61 (0.01) & 0.65 (0.01) \\
LIME & 0.50 (0.03) & 0.37 (0.12) & 0.42 (0.13) & 0.52 (0.05) & 0.33 (0.15) & 0.27 (0.06) \\
WISP & 0.31 (0.02) & 0.56 (0.01) & 0.47 (0.00) & 0.46 (0.01) & 0.53 (0.01) & 0.53 (0.02) \\
\bottomrule
\end{tabular}
\\[4pt]
\begin{tabular}{lccccc}
\toprule
Method & Aq.\ sol. & PPB plasma & PPB brain & Half-life & PPB muscle \\
\midrule
MolLedger & 0.25 (0.03) & 0.33 (0.02) & 0.33 (0.06) & 0.38 (0.00) & 0.33 (0.07) \\
MolLedg. no ctx & 0.20 (0.02) & 0.16 (0.01) & 0.13 (0.01) & 0.17 (0.03) & 0.17 (0.02) \\
Unanch MolLedg. & 0.27 (0.01) & 0.31 (0.07) & 0.35 (0.04) & 0.40 (0.02) & 0.27 (0.03) \\
Anch GNAN & 0.06 (0.00) & \textbf{0.06 (0.01)} & \textbf{0.03 (0.00)} & \textbf{0.06 (0.03)} & \textbf{0.04 (0.01)} \\
Unanch GNAN & \textbf{0.05 (0.00)} & \textbf{0.06 (0.01)} & 0.07 (0.06) & 0.08 (0.03) & 0.21 (0.13) \\
IG & 0.16 (0.04) & 0.34 (0.11) & 0.30 (0.08) & 0.33 (0.03) & 0.37 (0.10) \\
Grad-CAM & 0.42 (0.03) & 0.67 (0.01) & 0.62 (0.02) & 0.50 (0.10) & 0.69 (0.06) \\
LIME & 0.27 (0.02) & 0.39 (0.12) & 0.37 (0.09) & 0.48 (0.11) & 0.46 (0.10) \\
WISP & 0.35 (0.01) & 0.43 (0.04) & 0.47 (0.03) & 0.42 (0.06) & 0.55 (0.05) \\
\bottomrule
\end{tabular}
\end{table}

\begin{table}
\centering
\small
\caption{Explained $\Delta$ MAE. These are the numbers in the bottom rows of Figures~\ref{fig:interpretability} and~\ref{fig:interpretability_other}. Best shown in bold.}
\label{tab:interp-explained}
\vspace{0.2cm}
\begin{tabular}{lcccccc}
\toprule
Method & LogD & Kin.\ sol. & CL$_{int}$ mouse & CL$_{int}$ human & Caco2 P$_{app}$ & Caco2 efflux \\
\midrule
MolLedger & \textbf{0.38 (0.00)} & \textbf{0.96 (0.01)} & 1.04 (0.02) & 0.91 (0.04) & 0.42 (0.01) & \textbf{0.73 (0.01)} \\
MolLedg. no ctx & \textbf{0.38 (0.01)} & 0.98 (0.04) & \textbf{1.01 (0.01)} & 0.91 (0.02) & 0.44 (0.00) & 0.78 (0.02) \\
Unanch MolLedg. & 0.42 (0.02) & 0.98 (0.01) & 1.10 (0.03) & \textbf{0.90 (0.01)} & \textbf{0.38 (0.02)} & 0.75 (0.02) \\
Anch GNAN & 0.66 (0.00) & 1.07 (0.01) & 1.28 (0.00) & 1.09 (0.01) & 0.56 (0.00) & 0.94 (0.01) \\
Unanch GNAN & 0.66 (0.00) & 1.07 (0.02) & 1.28 (0.00) & 1.09 (0.01) & 0.55 (0.00) & 0.93 (0.01) \\
IG & 0.98 (0.39) & 2.28 (0.58) & 3.03 (1.92) & 1.57 (0.11) & 3.72 (3.05) & 3.54 (1.88) \\
Grad-CAM & 0.89 (0.03) & 1.29 (0.03) & 1.63 (0.05) & 1.52 (0.03) & 0.67 (0.02) & 1.13 (0.01) \\
LIME & 1.00 (0.28) & 2.42 (0.68) & 2.65 (1.74) & 1.45 (0.19) & 3.18 (2.56) & 3.22 (1.47) \\
WISP & 2.85 (0.21) & 4.49 (0.09) & 6.13 (0.23) & 5.41 (0.29) & 2.03 (0.03) & 3.77 (0.14) \\
\bottomrule
\end{tabular}
\\[4pt]
\begin{tabular}{lccccc}
\toprule
Method & Aq.\ sol. & PPB plasma & PPB brain & Half-life & PPB muscle \\
\midrule
MolLedger & 0.56 (0.05) & 0.60 (0.02) & \textbf{0.62 (0.03)} & \textbf{0.68 (0.10)} & 1.75 (0.08) \\
MolLedg. no ctx & \textbf{0.53 (0.02)} & \textbf{0.54 (0.03)} & \textbf{0.62 (0.01)} & 0.76 (0.02) & 2.30 (0.01) \\
Unanch MolLedg. & 0.59 (0.01) & 0.62 (0.04) & \textbf{0.62 (0.05)} & 0.81 (0.06) & 1.77 (0.05) \\
Anch GNAN & 1.10 (0.00) & 0.88 (0.03) & 0.78 (0.01) & 0.92 (0.02) & \textbf{1.47 (0.01)} \\
Unanch GNAN & 1.07 (0.00) & 0.88 (0.03) & 0.79 (0.01) & 0.91 (0.02) & 1.57 (0.05) \\
IG & 4.01 (0.49) & 2.22 (1.12) & 2.65 (1.44) & 1.85 (0.64) & 2.61 (0.59) \\
Grad-CAM & 1.00 (0.04) & 1.11 (0.05) & 0.97 (0.01) & 0.79 (0.06) & \textbf{1.47 (0.08)} \\
LIME & 3.60 (0.49) & 2.43 (1.29) & 2.44 (1.43) & 1.89 (0.59) & 2.67 (0.52) \\
WISP & 2.80 (0.18) & 3.51 (0.17) & 3.26 (0.26) & 2.89 (0.55) & 7.77 (0.67) \\
\bottomrule
\end{tabular}
\end{table}

\end{document}